\documentclass[10pt,twocolumn,letterpaper]{article}

\usepackage{iccv}
\usepackage{times}
\usepackage{epsfig}
\usepackage{graphicx}
\usepackage{amsmath}
\usepackage{amssymb}
\usepackage{enumitem}
\usepackage{multirow}
\usepackage{makecell}
\usepackage{graphicx}
\usepackage[table]{xcolor}
\usepackage{arydshln}
\usepackage{array} 
\usepackage{caption}
\usepackage{subcaption}
\usepackage{pifont}
\usepackage{bbding}

\usepackage[breaklinks=true,bookmarks=false]{hyperref}

\iccvfinalcopy 

\def\iccvPaperID{****} 

\begin{document}

\title{Towards Robust and Realistic Human Pose Estimation via WiFi Signals}

\author{%
  Yang Chen\textsuperscript{1}, Jingcai Guo\textsuperscript{1,2 \thanks{Jingcai Guo is the corresponding author.}}\\
\textsuperscript{1}Department of Computing, The Hong Kong Polytechnic University~\\
\textsuperscript{2}Department of Land Surveying and Geo-Informatics, The Hong Kong Polytechnic University~\\
\texttt{jc-jingcai.guo@polyu.edu.hk}
}

\maketitle

\begin{abstract}
Robust WiFi-based human pose estimation is a challenging task that bridges discrete and subtle WiFi signals to human skeletons. 
%
This paper revisits this problem and reveals two critical yet overlooked issues: 
1) cross-domain gap, i.e., due to significant variations between source-target domain pose distributions; and 2) structural fidelity gap, i.e., predicted skeletal poses manifest distorted topology, usually with misplaced joints and disproportionate bone lengths. 
This paper fills these gaps by reformulating the task 
into a novel two-phase framework dubbed \textit{\textbf{DT-Pose}}: \underline{\textit{\textbf{D}}}omain-consistent representation learning and \underline{\textit{\textbf{T}}}opology-constrained \underline{\textit{\textbf{Pose}}} decoding. 
Concretely, we first propose a temporal-consistent contrastive learning strategy with uniformity regularization, coupled with self-supervised masking-reconstruction operations, to enable robust learning of domain-consistent and motion-discriminative WiFi-specific representations. 
Beyond this, we introduce a simple yet effective pose decoder with task prompts, which integrates Graph Convolution Network (GCN) and Transformer layers to constrain the topology structure of the generated skeleton by exploring the adjacent-overarching relationships among human joints. 
Extensive experiments conducted on various benchmark datasets highlight the superior performance of our method in tackling these fundamental challenges in both 2D/3D human pose estimation tasks~\footnote{Code is available at: \url{https://github.com/cseeyangchen/DT-Pose}.}.
\end{abstract}    
\section{Introduction}
Image-based human pose estimation, a highly active and hot topic, has recently achieved remarkable success in both 2D \cite{cao2017realtime,wang2022lite} and 3D scenarios \cite{li2022exploiting,gong2023diffpose}, spanning single-person \cite{zhang2021single} and multi-person settings \cite{shi2022end,liu2023group}. These advancements have significantly propelled broad applications in virtual reality \cite{zheng2023deep}, autonomous driving \cite{zheng2022multi}, and the healthcare community \cite{he2024expert}. However, those visual-based methods face inherent limitations due to realistic challenges (e.g., lighting intensity, view variations, and occlusions). Furthermore, rising concerns regarding privacy have driven the growing research attention toward non-visual modalities (e.g., WiFi \cite{yan2024person,d2025hpe}, RF \cite{fan2025diffusion}, and wearable sensor \cite{chen2020survey} signals), which offer significant advantages in privacy protection and resilience to occlusions. Among these, the WiFi modality holds promise due to its widespread deployment and easy accessibility in the IoT era. 

\begin{figure*}
\begin{center}
\begin{tabular}{cccc}
    \hspace{-5pt}\includegraphics[width=0.32\linewidth, height=0.195\linewidth]{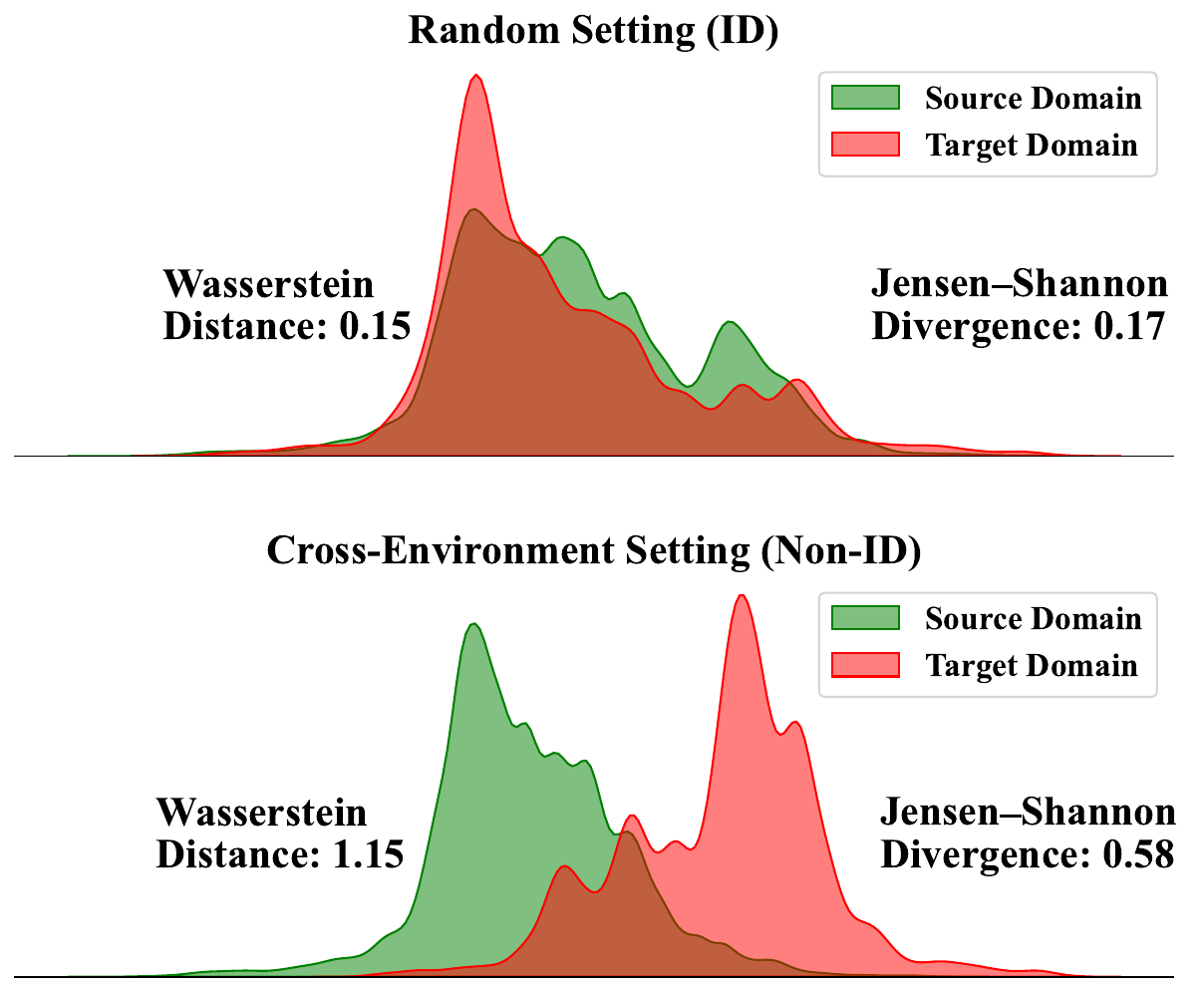} &
    \hspace{-5pt}\includegraphics[width=0.32\linewidth, height=0.195\linewidth]{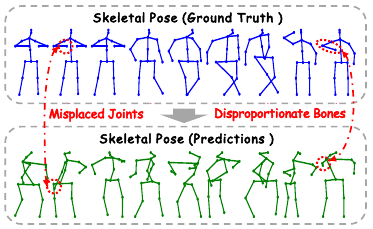} &
    \hspace{-5pt}\includegraphics[width=0.32\linewidth, height=0.195\linewidth]{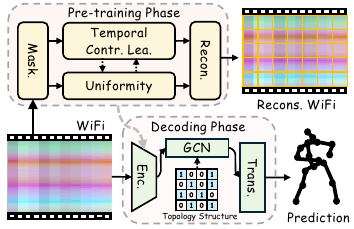}\\
    \hspace{-5pt}\small{(a) Pose Domain Gap} & 
    \hspace{-5pt}\small{(b) Distorted Predictions} &
    \hspace{-5pt}\small{(c) Two-Stage Process}\\
\end{tabular}
\end{center}
\vspace{-15pt}
\caption{(a) shows the pose coordinates distribution between the source and target domains. (b) represents the predictions of the MetaFi++ method \cite{zhou2023metafi++} and corresponding ground truth. (c) denotes the overview framework of our method.}
\label{fig:challenges}
\vspace{-15pt}
\end{figure*}

Tracing the development of the WiFi-based human pose estimation field, it has gradually evolved from single-person 2D to more complex multi-person 3D pose estimation \cite{wang2019can,wang2019person,jiang2020towards,ren2022gopose,zhou2022perunet,yang2022autofi,ren2021winect,zhou2023metafi++,yan2024person,d2025hpe}. Predominately, these methods rely on complex regression networks designed to map the WiFi signals to corresponding 2D/3D pose coordinates under the supervised learning paradigm. However, these approaches typically hypothesize an identical distribution of training and testing data, inconsistent with the variability encountered in real-world scenarios. To address this limitation, the recently introduced WiFi dataset (MM-Fi \cite{yang2024mm}) incorporates cross-domain settings, presenting new challenges for evaluating the generalizability of WiFi-based human pose estimation methods. 

Upon analyzing WiFi signals across diverse domains, we observed a notable difference in pose coordinates distribution between the source domain and target domain in the cross-environment setting, contrasting with the identical distribution hypothesis before (Fig. \ref{fig:challenges} (a)). Prior methods learned the task-specific representations in this setting that overfitted to pose distributions within the source domain while failing to generalize effectively to the target domain. This limitation indicates the inadequacy of the supervised learning paradigm in capturing the intrinsic motion properties of WiFi signals, even leading to the learning of spurious, motion-unrelated, and noise features that bias results toward the source domain. While the recently proposed AdaPose \cite{zhou2024adapose} model similarly notes this issue, its reliance on pre-acquired data from the target domain as a prerequisite for domain adaption renders it impractical and suboptimal. Thus, we aim to explore domain-consistent and motion-discriminative WiFi representations independent of pose coordinate space, thereby enhancing cross-domain transferability.

In addition to domain generalization challenges, we also observe that existing methods frequently predict distorted poses with unrealistic topologies (e.g., misplaced joints and disproportionate bone lengths), as shown in Fig. \ref{fig:challenges} (b). These deficiencies stem from the undesirable modeling of joint relationships during the pose decoding phase, which neglects the intricate spatial properties of human joints. Moreover, mapping joint features to pose coordinates without structural constraints proves inferior. To address these issues, we propose incorporating explicit skeletal structure constraints to better model the non-trivial spatial relationships among joints.

Building on the above observations, we propose a novel framework (\textit{\textbf{DP-Pose}}), which reformulates WiFi-based human pose estimation as a two-phase process: \textit{\textbf{D}}omain-consistent WiFi representation learning and \textit{\textbf{T}}opology-constrained \textit{\textbf{Pose}} decoding, as depicted in Fig. \ref{fig:challenges} (c). In the first phase, we initially transform the WiFi signals into image-like data and adopt the self-supervised masking-reconstruction operation in MAE \cite{he2022masked} as the main line to learn domain-consistent WiFi-specific representations. Considering the temporal property of WiFi signals, we introduce contrastive learning by treating adjacent WiFi frames within an action sequence as positive pairs and other batch WiFi samples as negative pairs, yielding more motion-discriminative representations. Additionally, uniformity regularization is applied to prevent dimensional collapse caused by the sparsity of WiFi signals. In the second phase, the pre-trained encoder is frozen to extract domain-consistent WiFi representations, followed by the introduction of task prompt and Graph Convolution Network (GCN) layers that leverage pre-defined human topology constraints to model local correlations among adjacent joints. Concurrently, we integrate the Transformer layers to establish more holistic dependencies among overarching joints. By exploring these adjacent-overarching spatial relationships of human joints, our framework ensures the generation of realistic poses that adhere to human skeletal topology. 

The main contributions can be summarized as follows:
\begin{itemize}
\item We develop a novel self-supervised method to learn domain-consistent and motion-discriminative WiFi representations using temporal contrastive masking-reconstruction strategy with uniformity regularization, effectively mitigating cross-domain gap challenges.
\item We propose a topology-constrained pose decoding mechanism that combines task prompt, Graph Convolution Network (GCN), and Transformer layers to generate realistic poses by exploring the intrinsic adjacent-overarching spatial characteristics of human joints.
\item We evaluate the superior performance of our method through extensive experiments on the MM-Fi, WiPose, and Person-in-WiFi-3D datasets, demonstrating its effectiveness in mitigating cross-domain and structural fidelity gaps for WiFi-based pose estimation tasks.
\end{itemize}

\begin{figure*}
\begin{center}
\includegraphics[width=\linewidth, height=0.4\linewidth]{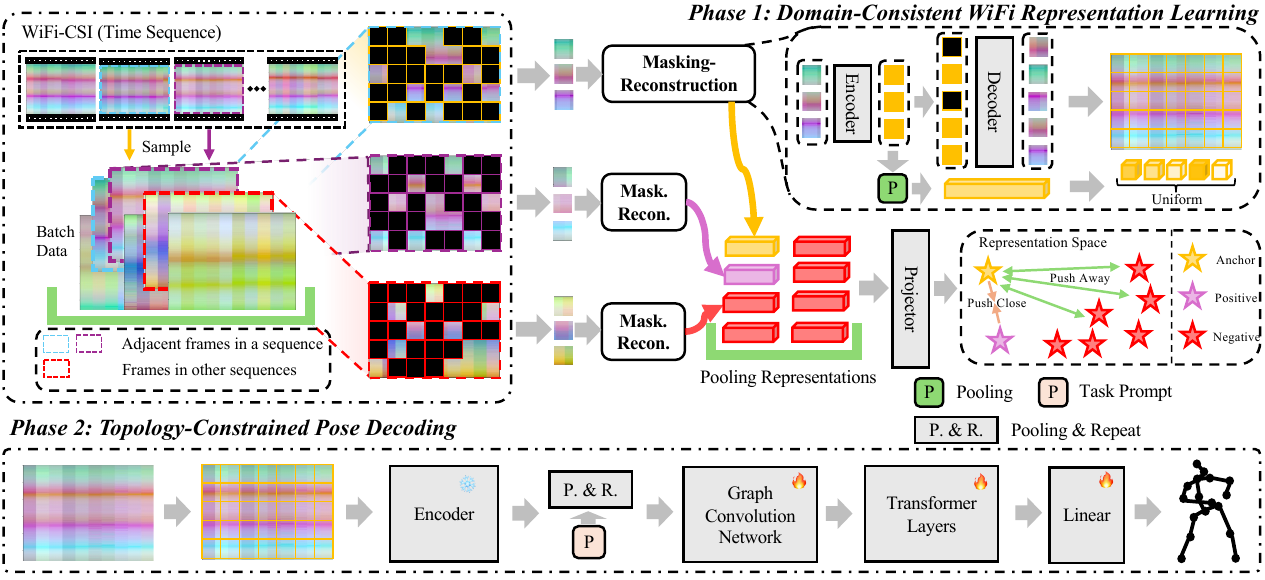} 
\end{center}
\vspace{-15pt}
\caption{The pipeline of our method, including the pre-training and pose decoding phases. 
}
\label{fig:framework}
\vspace{-15pt}
\end{figure*}

\section{Related Work}
\subsection{Masked Pre-training}
Masked pretraining techniques have been widely studied across various data modalities for self-supervised representation learning, leveraging the reconstruction of masked inputs as a core strategy \cite{devlin2018bert,radford2018improving,bao2021beit,he2022masked,tong2022videomae,wang2023videomae,huang2022masked,yan2023skeletonmae,cheng2023timemae}. Among these modalities, the BERT \cite{devlin2018bert} and GPT \cite{radford2018improving} are two seminal language models that pioneered the masked modeling paradigm by predicting masked word tokens based on context information. Inspired by them, the computer vision community introduced masked pertaining frameworks for images, giving rise to representative methods like BEiT \cite{bao2021beit} and MAE \cite{he2022masked}, while the video community \cite{tong2022videomae, wang2023videomae} subsequently demonstrated that the masked mechanism extends effectively into the temporal dimension.
Beyond these, other data modalities, including audio (Audio-MAE \cite{huang2022masked}), skeleton (SkeletonMAE \cite{yan2023skeletonmae}), and time series (TimeMAE \cite{cheng2023timemae}), have similarly validated the feasibility and efficacy of masked modeling for task-agnostic representation learning in a self-supervised manner.
[\textit{\textbf{Summary}}]: To the best knowledge of us, this work is the first to employ a self-supervised masked pre-training paradigm in the WiFi modality. Moreover, we propose the temporal-consistent contrastive strategy with uniformity regularization to obtain motion-discriminative representations.

\subsection{Skeleton-based Action Recognition}
Learning skeleton action representation can be conceptualized as the inverse process of human pose decoding. Typically, skeleton-based action recognition methods can be categorized into CNN-based, GCN-based, and Transformer-based \cite{wang2016action,chen2024fine,chen2021channel,song2022constructing,chi2022infogcn,chen2024neuron}. CNN-based methods transform skeleton sequences into image-like formats to extract discriminative representations \cite{wang2016action, liu2017enhanced}. 
In contrast, GCN-based methods model human joints and bones as graph nodes and edges, explicitly incorporating learnable adjacent matrix to explore spatial-temporal features, thus improving performance by a large margin \cite{chen2021channel,song2022constructing,chi2022infogcn,chen2024vision}. More recently, Transformer-based methods leverage self-attention mechanisms to capture long-range dependencies among joints \cite{plizzari2021spatial, gao2022focal,he2024enhancing}.
[\textit{\textbf{Summary}}]: Inspired by the superior performance of using GCN and Transformers to learn representations from the skeleton forwardly, we combine their advantages to decode skeletal poses from representations reversely, yielding more realistic and faithful pose prediction.

\subsection{WiFi-Based Pose Estimation}
WiFi-based human pose estimation is an emerging research topic that has gradually flourished in recent years, encompassing a range of tasks from single-person 2D \cite{wang2019can, d2025hpe} and 3D \cite{jiang2020towards,ren2022gopose,ren2021winect,zhou2023metafi++} to multi-person 2D \cite{wang2019person} and 3D scenarios \cite{yan2024person}. Early work in this field, such as WiSPPN \cite{wang2019can,wang2019person}, pioneers 2D pose estimation by employing fundamental CNN models \cite{he2016deep}. Subsequently, WiPose \cite{jiang2020towards} extends to 3D poses through a combination of CNN and RNN layers, thereby leveraging temporal dynamics to yield smoother skeletal predictions. Differently, both GoPose \cite{ren2022gopose} and Winect \cite{ren2021winect} methods leverage the two-dimensional angle-of-arrival features of WiFi signals to estimate 3D poses. 
More recent methods, such as MetaFi++ \cite{zhou2023metafi++} and Person-in-WiFi-3D \cite{yan2024person}, employ the Transformer layers to learn WiFi representations for single-/multi-person 3D pose estimation. 
Concurrently, HPE-Li \cite{d2025hpe} has designed dynamic kernels and integrated them into CNN layers to predict poses more efficiently. 
[\textit{\textbf{Summary}}]: In this work, we are the first to address the cross-domain gap challenge by the self-supervised contrastive masking-reconstruction paradigm. Simultaneously, we capture adjacent-overarching spatial correlations of joints to produce structural fidelity poses. 

\section{Method}

\begin{figure*}
\begin{center}
\begin{tabular}{ccccc}
    \includegraphics[width=0.17\linewidth, height=0.12\linewidth]{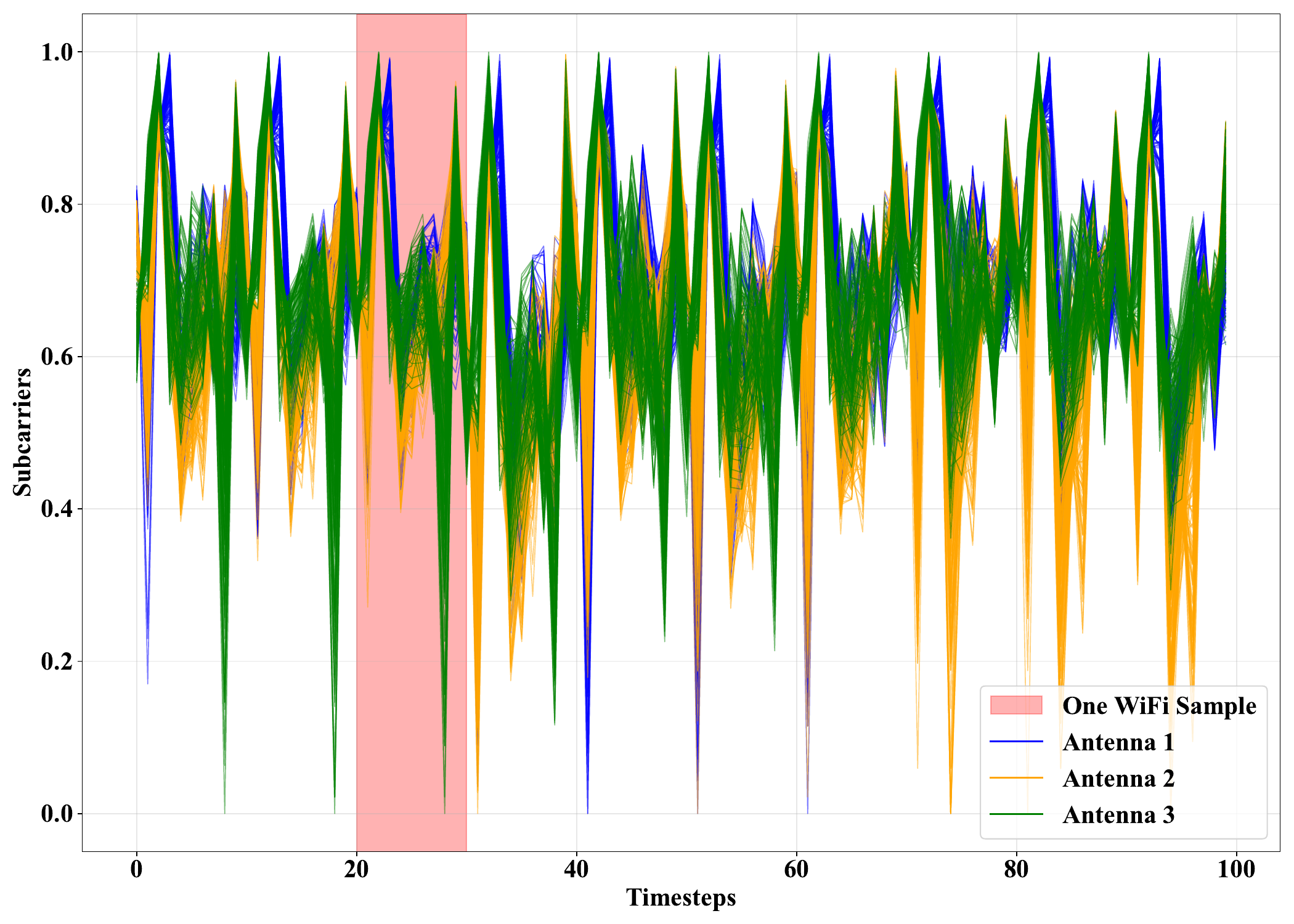} &
    \includegraphics[width=0.17\linewidth]{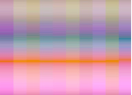} &
    \includegraphics[width=0.17\linewidth]{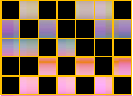} &
    \includegraphics[width=0.17\linewidth]{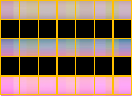} &
    \includegraphics[width=0.17\linewidth]{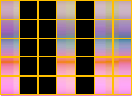} \\
    \small{(a) Signals} & 
    \small{(b) Image-like} &
    \small{(c) Unstructured} &
    \small{(d) Channel} &
    \small{(e) Time} \\
\end{tabular}
\end{center}
\vspace{-15pt}
\caption{Original WiFi CSI signals and different masking strategies on the MM-Fi dataset.}
\label{fig:wifi}
\vspace{-15pt}
\end{figure*}

\subsection{Preliminary}
\label{sec: preliminary}
Typically, WiFi signals are captured using multiple transmitters and receivers, with each signal comprising multiple subcarriers operating in orthogonal frequency bands to facilitate inter-device communication. These subcarriers describe the signal propagation process, known technically as Channel State Information (CSI). As shown in Fig. \ref{fig:wifi} (a), the CSI undergoes various distortions attributed to multipath effects and physical transformations, e.g., reflections, diffraction, and scattering \cite{yan2024person, zhou2023metafi++}. Leveraging these properties, we can record time-continuous WiFi signals that are dynamically influenced by human activities, i.e., action movements, thereby enabling the estimation of corresponding human poses. More specifically, one WiFi sample can be represented as $\mathbf{X} \in \mathbb{R}^{E\times R\times A\times S\times T}$, where $E, R, A, S$ denote the numbers of transmitters, receivers, antennas, and subcarriers, respectively. Here, $T=\frac{f_{\mathrm{wifi} }}{f_{\mathrm{video}}}$ represents the temporal resolution, equal to the ratio of the WiFi sampling frequency to that of the corresponding video action sequence. Notably, increasing the number of subcarriers and antennas enhances the resolution of the WiFi signals, capturing more subtle movements and finer variations. 
We define the ground truth of 3D pose coordinates for each frame as $\mathbf{Y}\in \mathbb{R}^{M\times J\times C}$, where $M$ represents the number of humans, $J$ indicates the number of joints, and $C$ specifies the spatial dimensions (coordinates). Hence, the entire dataset can be formalized as $\mathcal{D}=\{\mathbf{X}_{i}\in \mathbb{R}^{E\times R\times A\times S\times T}, \mathbf{Y}_{i}\in \mathbb{R}^{M\times J\times C}\}_{i=1}^{N}$, where $N$ is the total number of samples.

\subsection{Domain-Consistent Representation Learning}
\label{sec: phase1}

\noindent \textbf{Masking-Reconstruction Operation.}
To more closely align the WiFi modality with the image-based framework employed in MAE \cite{he2022masked}, we first reshape each WiFi sample into an image-like form $\mathbf{\hat{X}}_{i}\in \mathbb{R}^{A\times ERS \times T}$, as illustrated in Fig. \ref{fig:wifi} (b). Concretely, we treat the antenna dimension $A$ as the image channels, the concatenated subcarriers from all devices ($ERS$) to image height, and the temporal resolution $T$ as image width. Subsequently, to investigate the best suitable masking strategy tailored to WiFi signals, we consider three distinct approaches at the pre-training stage, including unstructured (i.e., random masking with grid shape), channel-structured (i.e., random masking subcarriers along the time-frequency axis), time-structured (i.e., random masking timesteps along the subcarrier axis), as shown in Fig. \ref{fig:wifi} (c) - (e). Following MAE \cite{he2022masked}, we divide $\mathbf{\hat{X}}_{i}$ into non-overlapped regular grid patches and employ convolution layers to embed each patch, obtaining $\mathbf{\tilde{X}}_{i}\in \mathbb{R}^{n\times d}$, where $n$ is the patch numbers and $d$ is the embedding dimension. We then incorporate fixed sinusoidal positional embeddings into these embedded patches and apply random masking with a high ratio (80\% in our experiments) to enforce robust representation learning. The encoder, comprising 4 stacked Transformer layers, is tasked with learning domain-consistent WiFi representations, while 2 Transformer layers in the decoder strive to reconstruct the original WiFi $\mathbf{\hat{X}}_{i}$, ultimately producing a reconstruction $\mathbf{X}'_{i}$. The entire masking-reconstruction procedure is optimized by minimizing the mean squared error (MSE) between the reconstruction and the original input as follows:
\begin{equation}
    \mathcal{L}_{\mathrm{Mask}} = ||\mathbf{X}'_{i} - \mathbf{\hat{X}}_{i}||_{2}^{2}.
\end{equation}

\noindent \textbf{Temporal-Consistent Contrastive Learning.}
Continuous WiFi samples captured over the duration of an action sequence can reflect the temporal variation of the motion. However, masking-reconstruction operation for individual WiFi samples would only extract modality-specific representations, lacking the essential motion patterns, as shown in Fig. \ref{fig:tsne-tc-cl} (a). Motivated by this, we treat adjacent WiFi frames within the same action sequence as a positive pair due to motion consistency within them. Notice that the WiFi samples in a batch include one-pair adjacent WiFi frames $(\mathbf{\hat{X}}_{t}, \mathbf{\hat{X}}_{t+1})$ from the same sequence and other non-isomorphic WiFi frames $\{\mathbf{\hat{X}}_{i}\}_{i=1}^{\mathrm{B}-2}$ from all action sequences, where $\mathrm{B}$ is the batch size. Thus, other combinations of WiFi samples in a batch should be negative pairs. Following the masking-reconstruction stage, we apply a pooling operation to the visible embedding patches of each sample from the encoder to derive a representation $\mathbf{e}_i \in \mathbb{R}^{d}$. Next, we project them to obtain positive pair representations $(\mathbf{s}_{t}, \mathbf{s}_{t+1})$ and other sample representations $\{\mathbf{v}_{i}\}_{i=1}^{\mathrm{B}-2}$. Then, we pull the positive pair closer and push the negative pairs away in a batch based on InfoNCE loss as follows:
\begin{eqnarray}
\mathcal{L}_{\mathrm{CL}}\!=\!-\mathrm{log}\frac{\mathrm{\phi}(\mathrm{\rho}(\mathrm{\mathbf{s}}_{t},\mathrm{\mathbf{s}}_{t+1})/\tau )}{\mathrm{\phi}(\mathrm{\rho}(\mathrm{\mathbf{s}}_{t},\mathrm{\mathbf{s}}_{t+1})/\tau)+\sum \nolimits_{i=1}^{\mathrm{B}-2} \mathrm{\phi}(\mathrm{\rho}(\mathrm{\mathbf{s}}_{t},\mathrm{\mathbf{v}}_{i})/\tau)},
\end{eqnarray}
where $\mathrm{\rho}(\cdot )$ denotes the cosine similarity, $\mathrm{\phi}(\cdot )$ is the $\mathrm{exp}(\cdot )$ function, and $\tau$ is the temperature parameter.

\noindent \textbf{Objective with Uniformity Regularization.}
The masking-reconstruction process encourages the extraction of WiFi-specific representations that are largely domain-agnostic, while the temporal-consistent contrastive learning objective ensures that these representations capture discriminative motion patterns. However, due to the inherent sparsity and homogeneity of WiFi signals, the learned representations may suffer from dimensional collapse, as depicted in Fig. \ref{fig:dim_collpase}. Here, we introduce explicit uniformity regularization to enhance representation diversity as follows:
\begin{equation}
    \mathcal{L}_{\mathrm{unif}}=\frac{1}{\mathrm{B}}\sum_{\substack{j=1, j \neq i}}^{\mathrm{B}}(\mathbf{e}_{i}^{\top}\mathbf{e}_{j})^2.
\end{equation}
Overall, the pre-training optimization objective for each WiFi sample can be formulated as follows:
\begin{equation}
    \mathcal{L}=\mathcal{L}_{\mathrm{Mask}} + \lambda_{\mathrm{CL}} \cdot 
 \mathcal{L}_{\mathrm{CL}} + \lambda_{\mathrm{unif}} \cdot \mathcal{L}_{\mathrm{unif}},
\end{equation}
where $\mathcal{L}_{\mathrm{CL}}$ and $\lambda_{\mathrm{unif}}$ are trade-off hyperparameters.

\subsection{Topology-Constrained Pose Estimation}
\label{sec: phase2}

\noindent \textbf{Adjacent Joint Local Modeling.}
After the pre-training phase, we freeze the WiFi encoder to extract WiFi representations $\mathbf{F}\in \mathbb{R}^{n\times d}$. To align these representations with the structure of human joints, we first pool all patches into one vector and repeat them into the number of human joints. Next, we add the learnable task prompt on them to obtain $\hat{\mathbf{F}} \in \mathbb{R}^{J\times d}$ for structural pose shape learning. 
Furthermore, we represent the human skeleton as a graph, where each joint is a vertex and each bone is an edge. This graph structure allows us to define the adjacency matrix $\mathbf{A}\in \{0, 1\}^{J\times J}$, where $\mathbf{A}_{i, j}=1$ indicates that the $i$-th joint and $j$-th joint are physical connected and $\mathbf{A}_{i, j}=0$ otherwise. By employing Graph Convolution Networks (GCN), we leverage adjacent matrix $\mathbf{A}$ to aggregate information from spatially connected joints. Formally, the updated representation $\tilde{\mathbf{F}}$ is computed as follows:
\begin{equation}
\tilde{\mathbf{F}}=\sigma(\mathbf{D}^{-\frac{1}{2}}\mathbf{A}\mathbf{D}^{-\frac{1}{2}}\mathbf{X}\mathbf{W}),
\end{equation}
where $\mathbf{D}\in \mathbb{R}^{J\times J}$ is the degree matrix for normalization, $\mathbf{W}$ is a learnable parameter, $\sigma$ is the activate function. 

\noindent \textbf{Overarching Joint Holistic Modeling.}
Beyond local relationships, it is essential to capture holistic, long-range correlations among overarching joints. To this end, we treat the joints as an ordered sequence and apply Transformer encoder layers to enhance their non-physical interdependencies, such as the potential relationships between head and hand joints in `drinking water' pose. We calculate the attention values among all joints as follows:
\begin{equation}
\mathbf{Q}=\mathbf{\tilde{\mathbf{F}}}\mathbf{W}_{Q}, \mathbf{K}=\mathbf{\tilde{\mathbf{F}}}\mathbf{W}_{K}, \mathbf{V}=\mathbf{\tilde{\mathbf{F}}}\mathbf{W}_{V}, 
\end{equation}
\begin{equation}
\mathbf{Z}_{\mathrm{attn}} =  \mathrm{LN}(\mathbf{\tilde{\mathbf{F}}}+\mathrm{softmax}(\frac{\mathbf{Q}\mathbf{K}^{T}}{\sqrt{\tilde{d}}})\mathbf{V}),
\end{equation}
where $\mathbf{W}_{Q}, \mathbf{W}_{K}, \mathbf{W}_{V}$ are learnable parameters, $\tilde{d}$ is the dimension of $\mathbf{K}$, and $\mathrm{LN}(\cdot)$ denotes the layer normalization. Then, we feed them into the feed-forward network $\mathrm{FFN}(\cdot)$ and regress them into pose coordinates by MLPs $\Psi (\cdot)$:
\begin{equation}
\mathbf{Z}= \mathrm{LN}(\mathrm{FFN}(\mathbf{Z}_{\mathrm{attn}})+\mathbf{Z}_{\mathrm{attn}}),
\quad \mathbf{\hat{Y}}_i = \Psi (\mathbf{Z}),
\end{equation}
where $\mathbf{\hat{Y}}_i$ is the predicted pose. By jointly capturing local dependencies among adjacent joints through GCNs and holistic relationships among overarching joints via Transformers, our method ensures that the predicted pose is structurally coherent and realistic.

\noindent \textbf{Objective.}
For training the pose decoder, we adopt the MSE loss for each sample to regress the pose as follows:
\begin{equation}
    \mathcal{L} = ||\mathbf{\hat{Y}}_i - \mathbf{Y}_i||_{2}^{2}.
\end{equation}

\section{Experiments}
\subsection{Dataset}
\noindent \textbf{MM-Fi} \cite{yang2024mm}.
It comprises 27 distinct action categories performed by 40 volunteers across four different rooms, resulting in approximately 320.76k single-person synchronized frames. 
One transmitter with one antenna and one receiver with three antennas capture all WiFi signals. Each skeletal pose consists of 17 joints encoded with 3D coordinates. To rigorously assess robustness, the dataset introduces three protocols and three settings for data splitting. Protocol 3 (P3) encompasses all 27 action categories, while Protocol 1 (P1) and Protocol 2 (P2) focus on 14 daily activities and 13 rehabilitation exercises, respectively. Setting 1 (S1 Random Split) randomly divides all data into training and testing sets with a 3:1 ratio. Setting 2 (S2 Cross-Subject Split) employs 32 subjects for training and the remaining 8 subjects for testing. Setting 3 (S3 Cross-Environment Split) selects 3 rooms randomly for training and others for testing.

\noindent \textbf{WiPose} \cite{zhou2022perunet}.
It contains 12 action categories performed by 12 volunteers. WiFi signals are captured by one transmitter with three antennas and one receiver with three antennas. Each pose annotation comprises 18 joints in 2D coordinates. The official split provides 132847 WiFi samples for training and 33753 for testing. 

\noindent \textbf{Person-in-WiFi-3D} \cite{yan2024person}.
It includes 8 daily actions performed by 7 volunteers at three distinct locations. 
A single transmitter with one antenna and three receivers with three antennas capture all WiFi signals. Each skeleton pose features 14 joints with 3D coordinates. The dataset has been officially partitioned into training and test sets, with 89946 WiFi samples allocated for training and 7824 for testing.

\begin{table*}
\caption{State-of-the-art comparisons on MM-Fi dataset regarding 3D pose estimation. The best and the second-best results are marked in \textcolor{red}{\textbf{Red}} and \textcolor{blue}
{\textbf{Blue}}, respectively. }
\vspace{-8pt}
\label{table:3d_mmfi}
\centering
\scalebox{0.66}{
\renewcommand{\arraystretch}{1.3}
\begin{tabular}{l|cccc|cccc|cccc}
\hline
\rule{0pt}{12pt} & \multicolumn{4}{c|}{Protocol 1} & \multicolumn{4}{c|}{Protocol 2} & \multicolumn{4}{c}{Protocol 3} \\
\hline
\rule{0pt}{12pt} Method & PCK@20$\uparrow$ & PCK@50$\uparrow$ & MPJPE$\downarrow$ & PA-MPJPE$\downarrow$ & PCK@20$\uparrow$ & PCK@50$\uparrow$ & MPJPE$\downarrow$ & PA-MPJPE$\downarrow$ & PCK@20$\uparrow$ & PCK@50$\uparrow$ & MPJPE$\downarrow$ & PA-MPJPE$\downarrow$ \\
\hline
\rowcolor{gray!10} \multicolumn{13}{l}{\textit{Setting 1 (Random Split)}:}\\
MetaFi++ \cite{zhou2023metafi++} & 49.1 & 86.5 & 186.9 & 120.7 & 32.2 & 81.7 & 213.5 & 121.4 & 43.9 & 85.0 & 197.1 & 121.2\\
HPE-Li \cite{d2025hpe} & \textcolor{blue}{\textbf{56.2}}  & \textcolor{blue}{\textbf{87.6}}  & \textcolor{blue}{\textbf{173.4}}  & \textcolor{blue}{\textbf{104.5}} &  \textcolor{blue}{\textbf{36.9}} & \textcolor{blue}{\textbf{81.9}}  & \textcolor{blue}{\textbf{206.1}}  & \textcolor{blue}{\textbf{102.7}} & \textcolor{blue}{\textbf{49.6}}  & \textcolor{blue}{\textbf{85.6}} & \textcolor{blue}{\textbf{184.3}} & \textcolor{blue}{\textbf{106.4}} \\
\rowcolor{yellow!10} \textbf{DT-Pose (Ours)} & \textcolor{red}{\textbf{59.4}}  & \textcolor{red}{\textbf{88.9}}  & \textcolor{red}{\textbf{165.3}}  & \textcolor{red}{\textbf{101.0}}  & \textcolor{red}{\textbf{41.4}}  & \textcolor{red}{\textbf{83.5}}  & \textcolor{red}{\textbf{195.6}}  &  \textcolor{red}{\textbf{101.2}} & \textcolor{red}{\textbf{51.7}}  & \textcolor{red}{\textbf{86.5}}  & \textcolor{red}{\textbf{178.5}}  &  \textcolor{red}{\textbf{104.5}} \\

\hdashline
\rowcolor{gray!10} \multicolumn{13}{l}{\textit{Setting 2 (Cross-Subject)}:} \\
MetaFi++ \cite{zhou2023metafi++} & 36.4 & \textcolor{blue}{\textbf{85.5}} & \textcolor{blue}{\textbf{222.3}} & 125.4 & 24.0 & 77.5 & 247.0 & 122.7 & 32.3 & \textcolor{blue}{\textbf{81.9}} & 231.1 & 124.0\\
HPE-Li \cite{d2025hpe} & \textcolor{blue}{\textbf{38.2}}  & 82.8 & 228.6  & \textcolor{blue}{\textbf{106.8}}  & \textcolor{blue}{\textbf{26.9}}  & \textcolor{blue}{\textbf{78.0}}  &  \textcolor{blue}{\textbf{242.6}} & \textcolor{blue}{\textbf{101.9}}  & \textcolor{blue}{\textbf{36.5}}  &  80.8 & \textcolor{blue}{\textbf{228.6}}  &  \textcolor{blue}{\textbf{107.7}} \\
\rowcolor{yellow!10}  \textbf{DT-Pose (Ours)} & \textcolor{red}{\textbf{41.9}}  & \textcolor{red}{\textbf{86.7}}  &  \textcolor{red}{\textbf{213.0}} & \textcolor{red}{\textbf{105.6}}  & \textcolor{red}{\textbf{28.5}}  & \textcolor{red}{\textbf{78.5}}  & \textcolor{red}{\textbf{238.3}}  &  \textcolor{red}{\textbf{101.1}} & \textcolor{red}{\textbf{37.7}}  & \textcolor{red}{\textbf{82.6}}  & \textcolor{red}{\textbf{221.6}} & \textcolor{red}{\textbf{106.2}}\\

\hdashline
\rowcolor{gray!10} \multicolumn{13}{l}{\textit{Setting 3 (Cross-Environment)}:} \\
MetaFi++ \cite{zhou2023metafi++} & \textcolor{blue}{\textbf{9.3}} & \textcolor{blue}{\textbf{55.1}} & \textcolor{blue}{\textbf{367.8}} & 121.0 & \textcolor{red}{\textbf{5.3}} & \textcolor{blue}{\textbf{45.9}} & \textcolor{blue}{\textbf{360.2}} & 117.2 & \textcolor{blue}{\textbf{6.4}} & \textcolor{blue}{\textbf{49.1}} & \textcolor{blue}{\textbf{369.5}} & 116.0 \\
HPE-Li \cite{d2025hpe} & 4.3  & 47.8  & 381.1  &  \textcolor{blue}{\textbf{110.3}} & 4.2  & 40.3  & 378.2  & \textcolor{blue}{\textbf{104.0}}  & 3.4  & 41.9  & 388.4  &  \textcolor{blue}{\textbf{107.9}} \\
\rowcolor{yellow!10} \textbf{DT-Pose (Ours)} & \textcolor{red}{\textbf{10.7}}  &  \textcolor{red}{\textbf{58.8}} & \textcolor{red}{\textbf{332.7}}  & \textcolor{red}{\textbf{105.1}}  & \textcolor{blue}{\textbf{4.4}}  & \textcolor{red}{\textbf{49.7}}  & \textcolor{red}{\textbf{338.3}}  & \textcolor{red}{\textbf{102.0}}  & \textcolor{red}{\textbf{9.8}}  & \textcolor{red}{\textbf{61.2}}  & \textcolor{red}{\textbf{316.8}}  &  \textcolor{red}{\textbf{104.2}} \\
\hline
\end{tabular}}
\end{table*}

\begin{table}[h]
\caption{State-of-the-art comparisons on the WiPose dataset regarding 2D pose estimation. The best and the second-best results are marked in \textcolor{red}{\textbf{Red}} and \textcolor{blue}{\textbf{Blue}}, respectively. }
\vspace{-8pt}
\label{table:2d_wipose}
\centering
\begin{tabular}{l|cc}
\hline
\rule{0pt}{10pt} Method & MPJPE$\downarrow$ & PA-MPJPE$\downarrow$ \\
\hline
MetaFi++ \cite{zhou2023metafi++}  & 49.2 & 30.1 \\
HPE-Li \cite{d2025hpe}  & \textcolor{blue}{\textbf{40.9}} & \textcolor{blue}{\textbf{25.9}} \\
\rowcolor{yellow!10} \textbf{DT-Pose (Ours)} &  \textcolor{red}{\textbf{34.3}} & \textcolor{red}{\textbf{23.1}} \\
\hline
\end{tabular}
\end{table}


\begin{table}
\caption{State-of-the-art comparisons on the Person-in-WiFi-3D (One-Person) dataset regarding 3D pose estimation. The best and the second-best results are marked in \textcolor{red}{\textbf{Red}} and \textcolor{blue}{\textbf{Blue}}, respectively. }
\vspace{-8pt}
\label{table:3d_piw}
\centering
\begin{tabular}{l|cc}
\hline
\rule{0pt}{10pt} Method & MPJPE$\downarrow$  & PA-MPJPE$\downarrow$ \\
\hline
MetaFi++ \cite{zhou2023metafi++} & 132.0 & 75.8 \\
HPE-Li \cite{d2025hpe} & 120.2 & \textcolor{blue}{\textbf{69.5}} \\
Wi-Pose \cite{jiang2020towards} & 101.8 & - \\
Person-in-WiFi-3D \cite{yan2024person} & \textcolor{blue}{\textbf{91.7}} & -  \\
\rowcolor{yellow!10} \textbf{DT-Pose (Ours)} & \textcolor{red}{\textbf{90.0}} & \textcolor{red}{\textbf{58.7}} \\
\hline
\end{tabular}
\end{table}

\subsection{Comparison with the State-of-the-Art Methods}
\noindent \textbf{3D Pose Estimation}. 
We evaluate our proposed method comprehensively for 3D pose estimation, comparing its performance against a range of state-of-the-art methods, as shown in Table \ref{table:3d_mmfi} and Table \ref{table:3d_piw}. Notably, our framework outperforms all existing methods across three metrics in each protocol and setting, demonstrating robust generalization capability. In particular, the superior PA-MPJPE results attest to the plausibility of our predicted poses, underscoring that the generated skeletal structures are topologically coherent. Moreover, the remarkable gains observed under the cross-domain settings confirm that our pre-training phase successfully captures general representations, while the subsequent pose decoder effectively establishes the correlations between representations and pose spaces.

\noindent \textbf{2D Pose Estimation}.
To further reveal the versatility of our self-supervised method, we evaluate its performance on 2D pose estimation tasks, as shown in Table \ref{table:2d_wipose} and Table \ref{table:2d_mmfi} (in Appendix). Compared to existing methods, our framework achieves the best results, showing the effectiveness of our design in learning task-agnostic WiFi representations.

\subsection{Ablation Study}
\noindent \textbf{Influence of Masking Strategies}.
In Table \ref{tab:as_masking}, we present a comparative analysis of masking strategies. The experimental results indicate that the unstructured strategy performs best during pre-training, primarily because it can attend to contextual cues spanning both time and channel levels.

\begin{table}
\caption{Masking strategies on the MM-Fi (P1-S1). }
\vspace{-8pt}
\centering
\begin{tabular}{c|cc}
\hline
Masking Strategy &  MPJPE$\downarrow$ & PA-MPJPE$\downarrow$ \\
\hline
Channel-Structured & 175.2 & 103.9 \\
Time-Structured & 180.7 & 107.1 \\
\hdashline
Unstructured (Ours) & 165.3 & 101.0\\
\hline
\end{tabular}
\label{tab:as_masking}
\end{table}

\begin{figure*}
\begin{center}
\begin{tabular}{ccccc}
    \includegraphics[width=0.22\linewidth, height=0.14\linewidth]{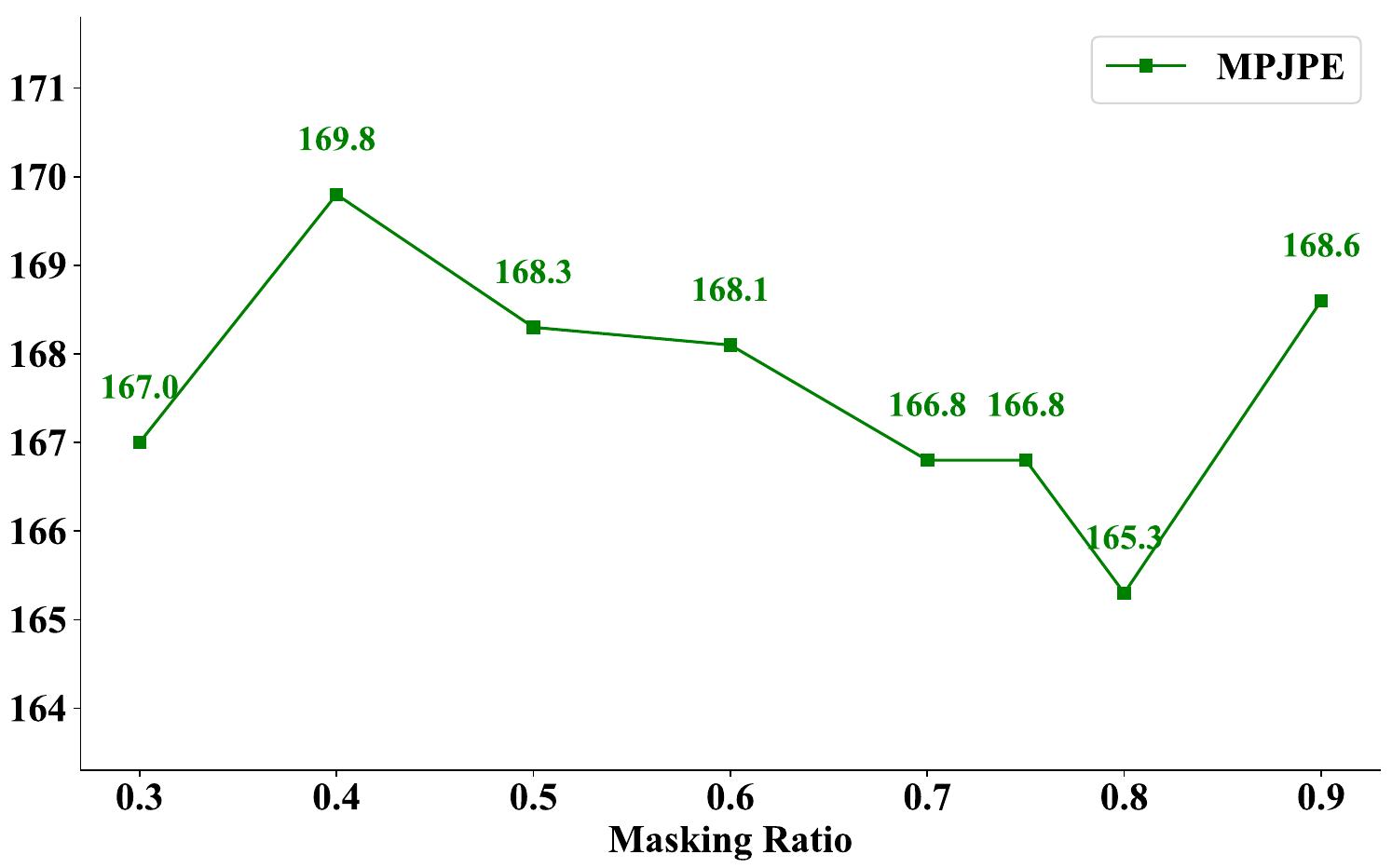} &
    \includegraphics[width=0.22\linewidth, height=0.14\linewidth]{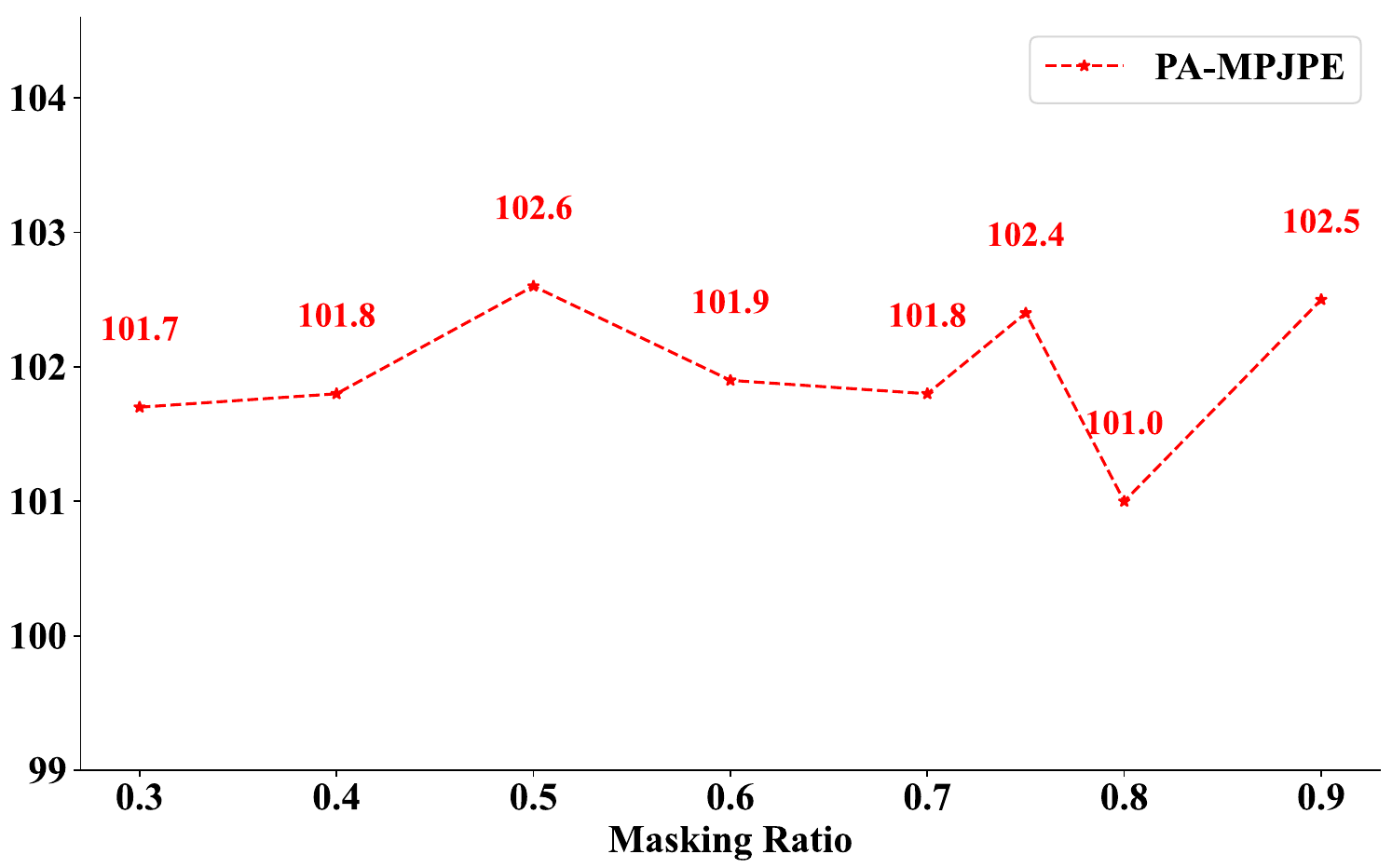} &
    \includegraphics[width=0.22\linewidth, height=0.14\linewidth]{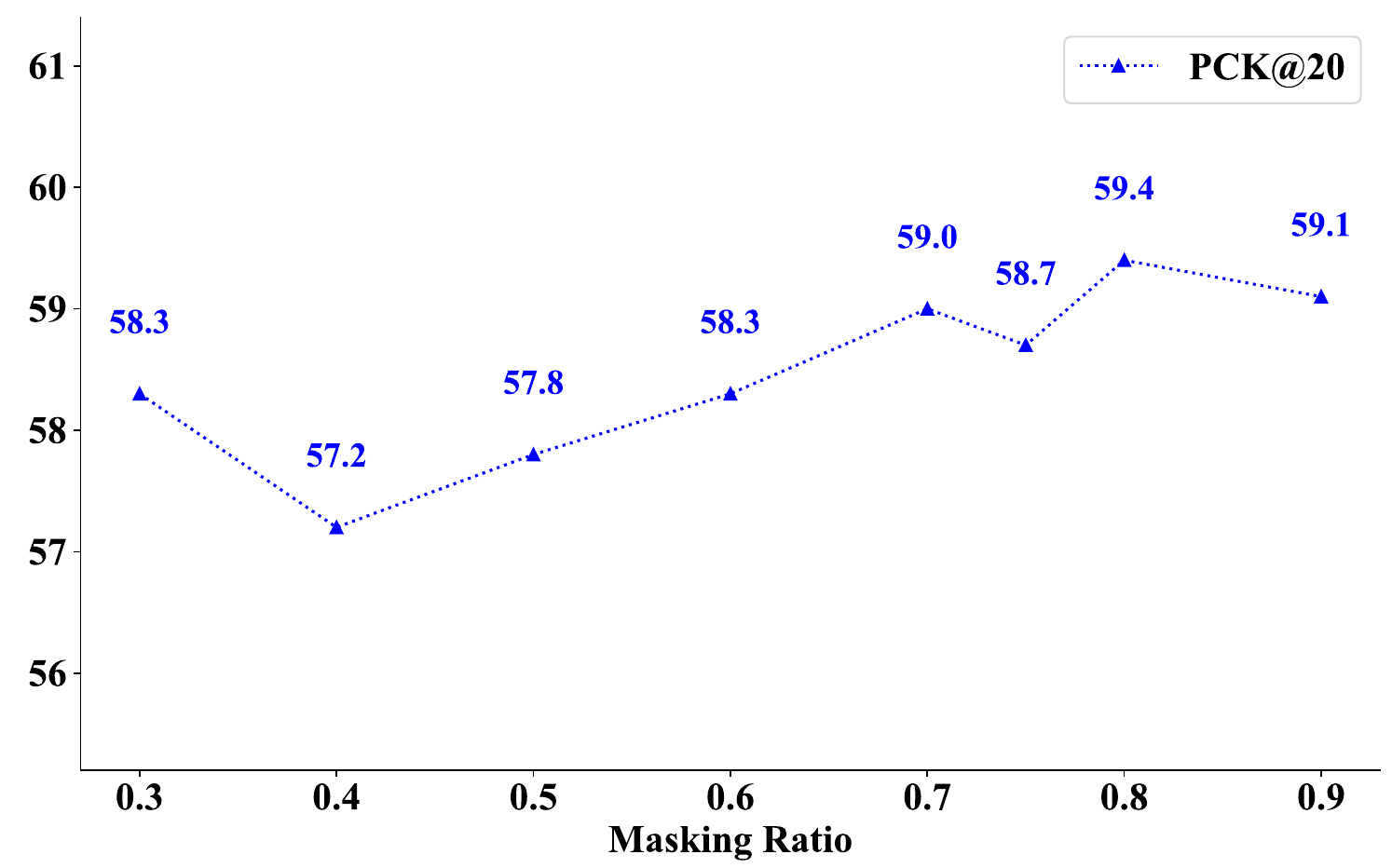} &
    \includegraphics[width=0.22\linewidth, height=0.14\linewidth]{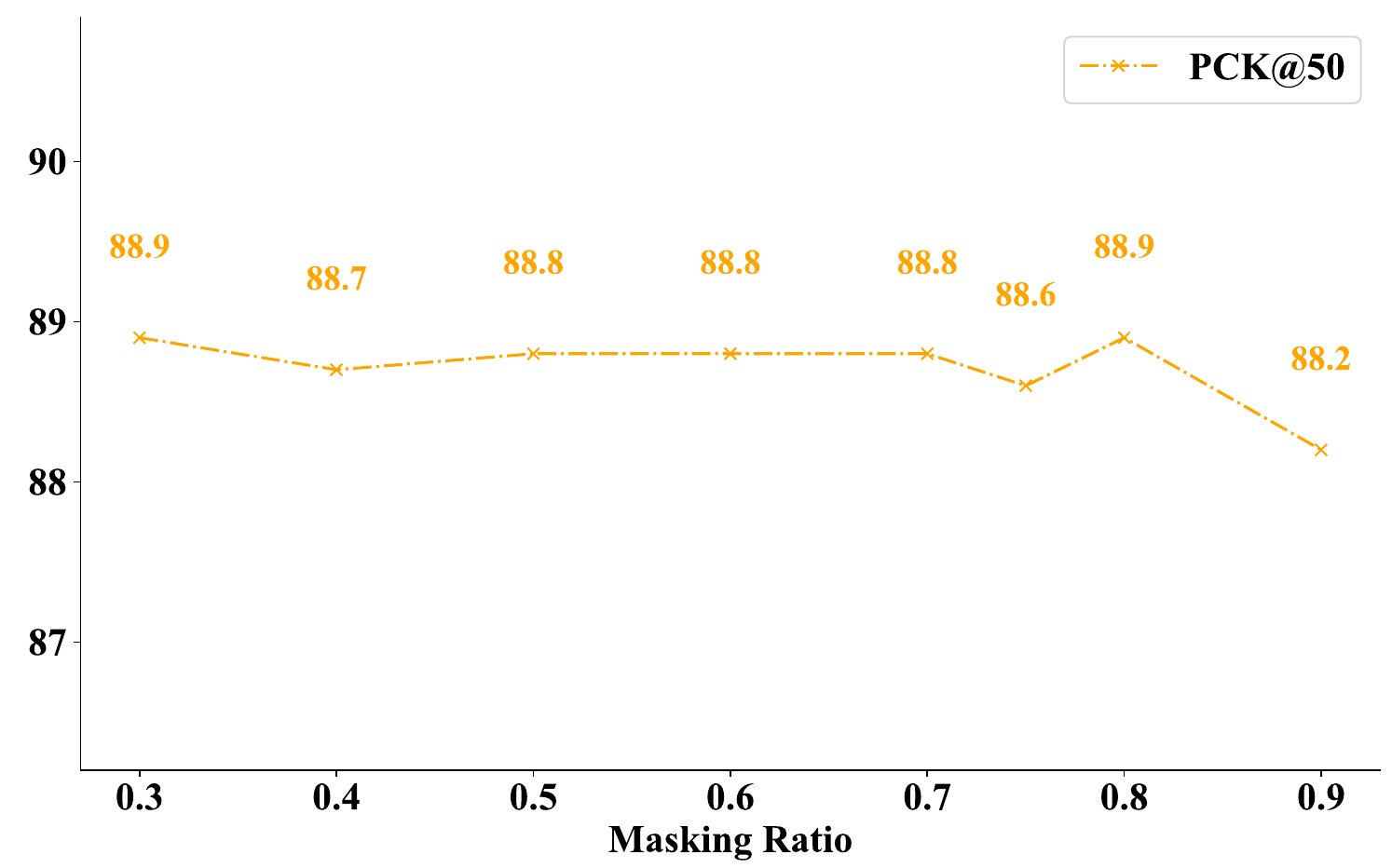}\\
\end{tabular}
\end{center}
\vspace{-15pt}
\caption{Performance on the MM-Fi (Protocol 1 - Setting 1) dataset with different masking ratios.}
\vspace{-10pt}
\label{fig:as_ratio}
\end{figure*}

\noindent \textbf{Influence of Masking Ratios}.
In Fig. \ref{fig:as_ratio}, we study the effect of varying masking ratios. The performance initially improves as the masking ratios increase but begins to decline once the ratio surpasses 80\%, likely due to the rising difficulty of reconstructing a large portion of inputs. Thus, we use the default masking ratio at 80\%.

\noindent \textbf{Influence of Pre-training Components}.
Table \ref{tab:as_phase1} studies the influence of each component in our pre-training phase. Notably, our per-training phase yields more general and robust representations than the model trained from scratch. Moreover, the temporal-consistent contrastive strategy contributes a lot, highlighting the necessity of capturing motion dynamics in WiFi-based pose estimation.

\begin{table}
\caption{Component analysis in the pre-training phase on the MM-Fi (P1-S1). MR: mask-reconstruction; TC-CL: temporal-consistent contrastive learning.}
\vspace{-8pt}
\centering
\begin{tabular}{ccc|cc}
\hline
MR & TC-CL & Uniformity & MPJPE$\downarrow$ & PA-MPJPE$\downarrow$ \\
\hline
\rowcolor{gray!10} \multicolumn{3}{l|}{\textit{Scratch}:} & & \\
\ding{55} & \ding{55} & \ding{55}& 198.6 & 100.9 \\
\hdashline
\ding{51} & \ding{55} & \ding{55}& 183.1 & 102.0 \\
\ding{51} & \ding{51} & \ding{55}& 173.1 & 102.7 \\
\ding{51} & \ding{55} & \ding{51}& 181.8 & 101.9 \\
\hdashline
\ding{51} & \ding{51} & \ding{51}& 165.3 & 101.0\\
\hline
\end{tabular}
\label{tab:as_phase1}
\end{table}

\begin{table}
\caption{Component analysis in the pose decoder on the MM-Fi (P1-S1). $\dagger$: MLPs as the pose decoder; $\ddagger$: we transform all patches into the number of joints by MLPs.}
\vspace{-8pt}
\centering
\scalebox{0.88}{
\begin{tabular}{ccc|cc}
\hline
\makecell[c]{Task\\Prompt} & GCN & Transformers & MPJPE$\downarrow$ & PA-MPJPE$\downarrow$ \\
\hline
\ding{55} & \ding{55} & \ding{55}& 197.4$^{\dagger}$ & 103.5$^{\dagger}$ \\
\ding{51} & \ding{55} & \ding{55}& 174.1$^{\phantom{\dagger}}$ & 101.3$^{\phantom{\dagger}}$ \\
\ding{55} & \ding{51} & \ding{55}& 179.8$^{\ddagger}$ & 107.0$^{\ddagger}$\\
\ding{55} & \ding{55} & \ding{51}& 181.4$^{\ddagger}$ & 103.0$^{\ddagger}$ \\
\ding{51} & \ding{51} & \ding{55}& 166.7$^{\phantom{\dagger}}$ & 103.2$^{\phantom{\dagger}}$ \\
\ding{51} & \ding{55} & \ding{51}& 167.1$^{\phantom{\dagger}}$ & 101.1$^{\phantom{\dagger}}$ \\
\ding{55} & \ding{51} & \ding{51}& 167.0$^{\phantom{\dagger}}$ & 103.3$^{\phantom{\dagger}}$ \\
\hdashline
\ding{51} & \ding{51} & \ding{51}& \textbf{165.3}$^{\phantom{\dagger}}$ & \textbf{101.0}$^{\phantom{\dagger}}$\\
\hline
\end{tabular}}
\label{tab:as_phase2}
\vspace{-5pt}
\end{table}

\noindent \textbf{Influence of Pose Decoding Components}.
We perform a detailed ablation study of each module within the pose decoder, as depicted in Table \ref{tab:as_phase2}. Notably, substituting simple MLP decoding with our adjacent-overarching joint modeling strategy significantly improves the accuracy of absolute joint localization, emphasizing the importance of capturing both local and global skeletal dependencies. Furthermore, the task prompt, which adaptive learns human skeletal priors into the decoding process, proves to be critical. If we remove it, the PA-MPJPE metrics will degrade a lot.

\begin{figure}
\begin{center}
\includegraphics[width=\linewidth]{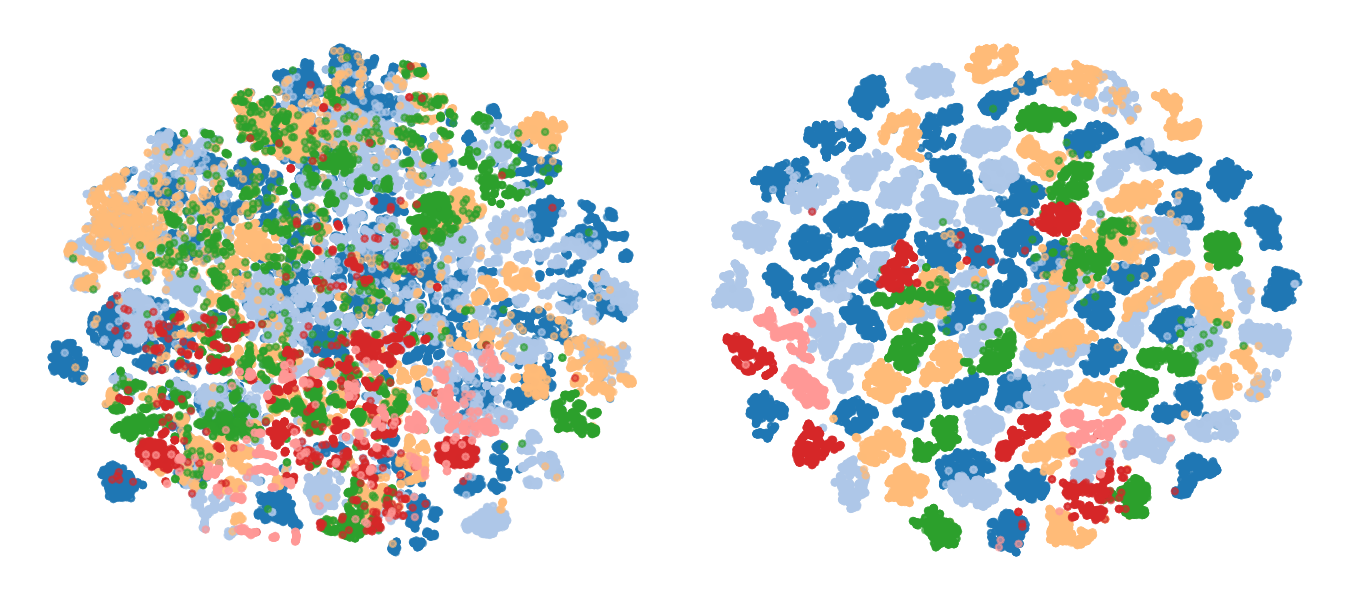}\\ \vspace{-2pt}
\begin{tabular}{@{\hskip 20pt}c@{\hskip 65pt}c@{\hskip 20pt}}
    \small{(a) w/o TC-CL} &
    \small{(b) w/ TC-CL} \\
\end{tabular}
\end{center}
\vspace{-15pt}
\caption{t-SNE visualization of WiFi representations on the MM-Fi (P1-S1). (a) represents that we drop the temporal-consistent contrastive strategy in the pre-training phase. (b) denotes that we equip it. Each color corresponds to a distinct action category.}
\label{fig:tsne-tc-cl}
\end{figure}

\begin{figure}
\begin{center}
\includegraphics[width=\linewidth, height=0.4\linewidth]{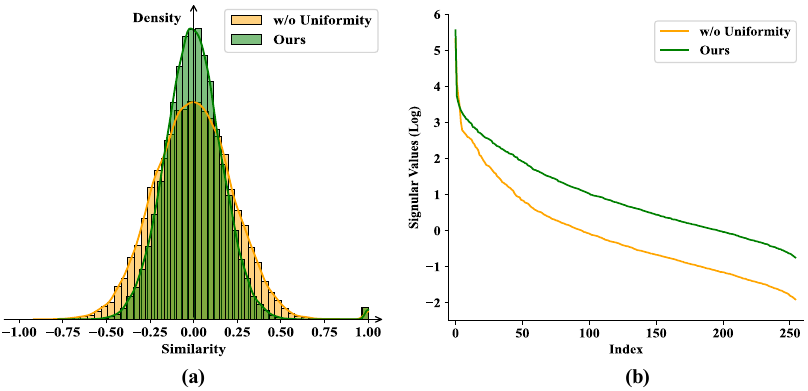} 
\end{center}
\vspace{-15pt}
\caption{Dimension collapse. (a) represents the statistics of the covariance values of the WiFi representation dimensions. (b) compares the singular values of WiFi representations (zoom in for a better view).}
\label{fig:dim_collpase}
\end{figure}

\begin{figure*}
\begin{center}
\begin{tabular}{cccc}
    \includegraphics[width=0.21\linewidth]{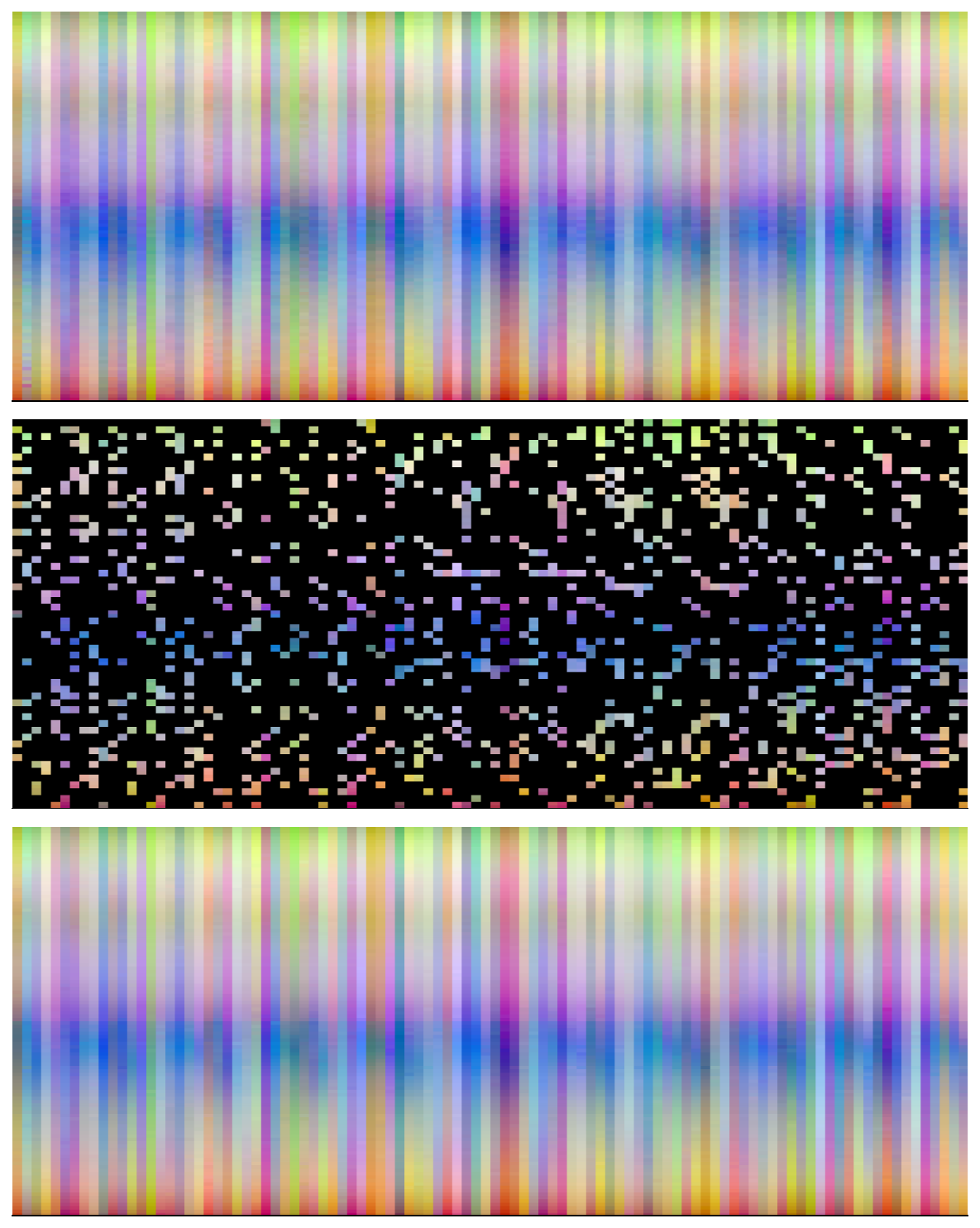} &
    \includegraphics[width=0.21\linewidth]{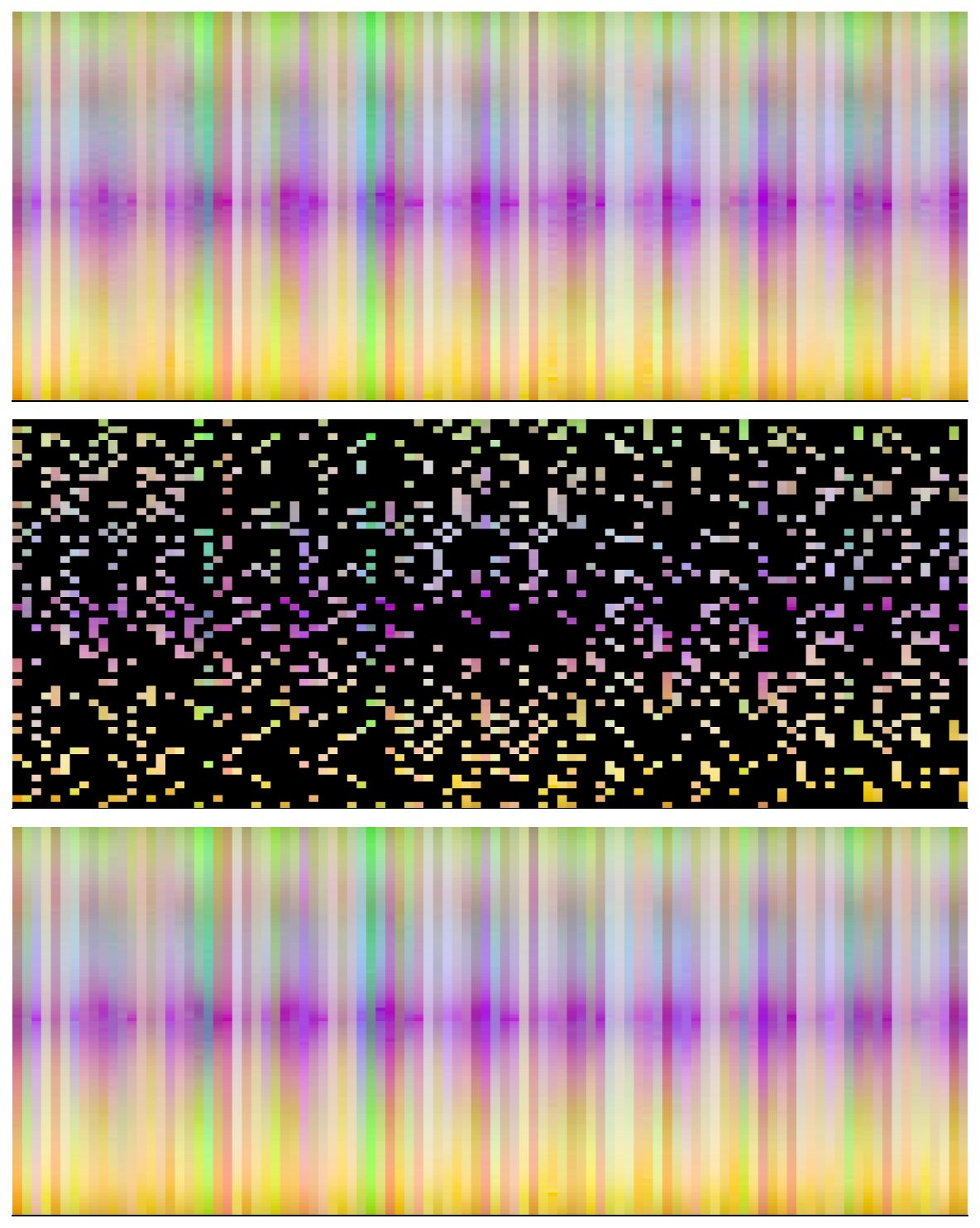} &
    \includegraphics[width=0.21\linewidth]{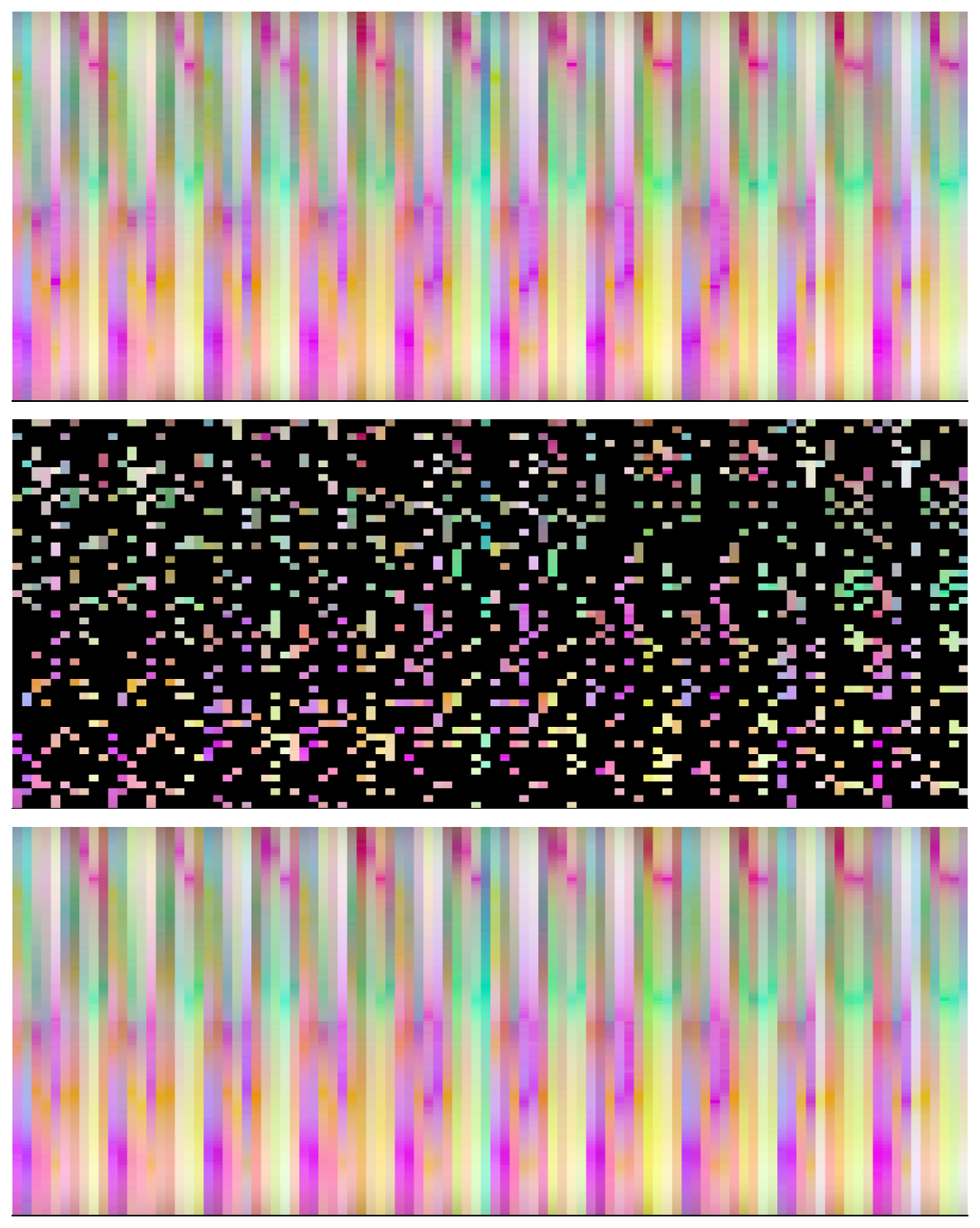} &
    \includegraphics[width=0.21\linewidth]{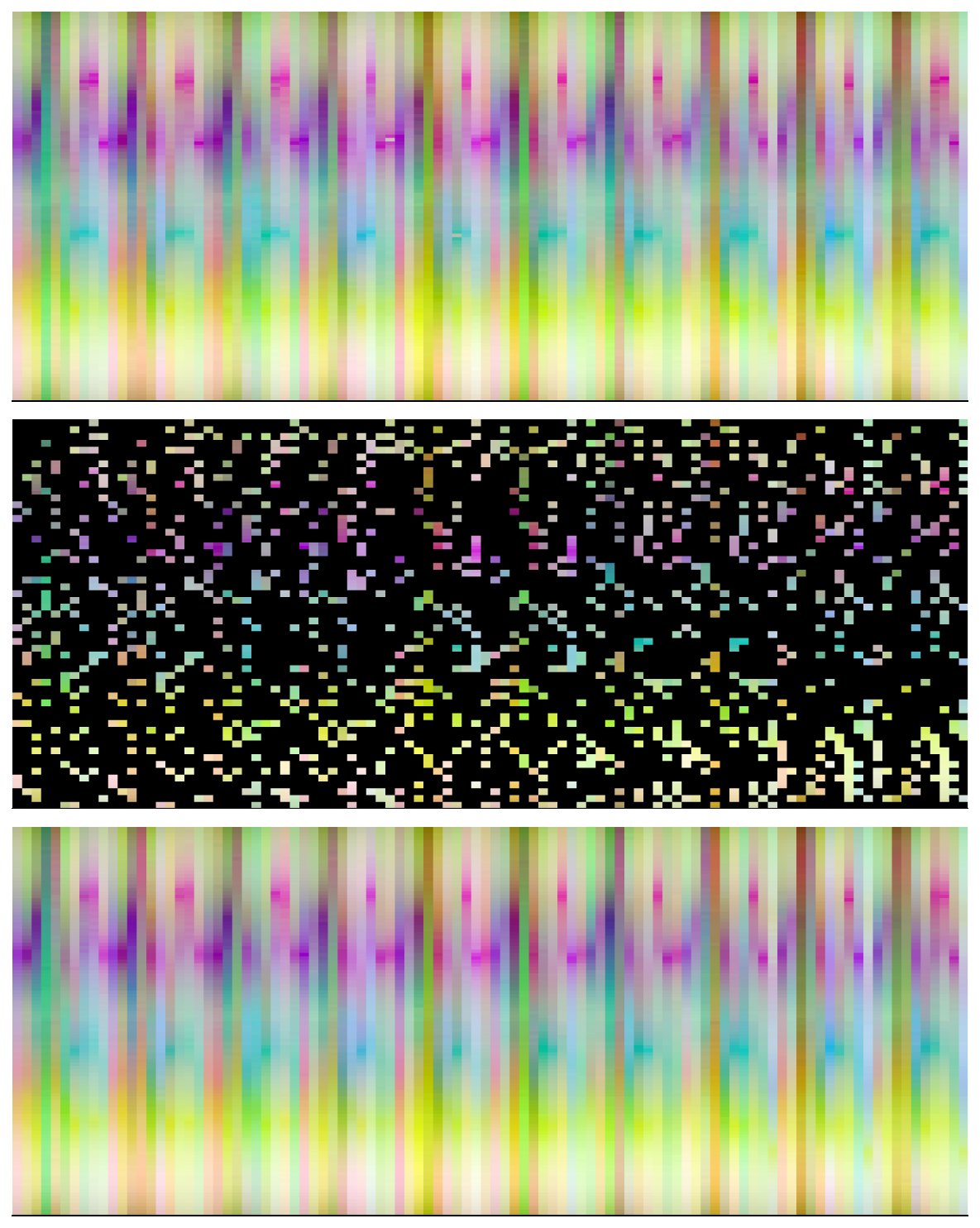} \\
    \small{(a) Chest Expanding Horizontally} & 
    \small{(b) Chest Expanding Vertically} &
    \small{(c) Raising Right Arm} &
    \small{(d) Waving Left Arm} \\
\end{tabular}
\end{center}
\vspace{-15pt}
\caption{WiFi visualizations on the MM-Fi (P3-S1). The first row represents the raw WiFi signals, the second row represents the masked WiFI input, and the third denotes the reconstructed WiFi output. All of them contain ten continuous frames.}
\label{fig:ori_mask_recon}
\vspace{-10pt}
\end{figure*}

\begin{figure*}
\begin{center}
\includegraphics[width=0.95\linewidth, height=0.38\linewidth]{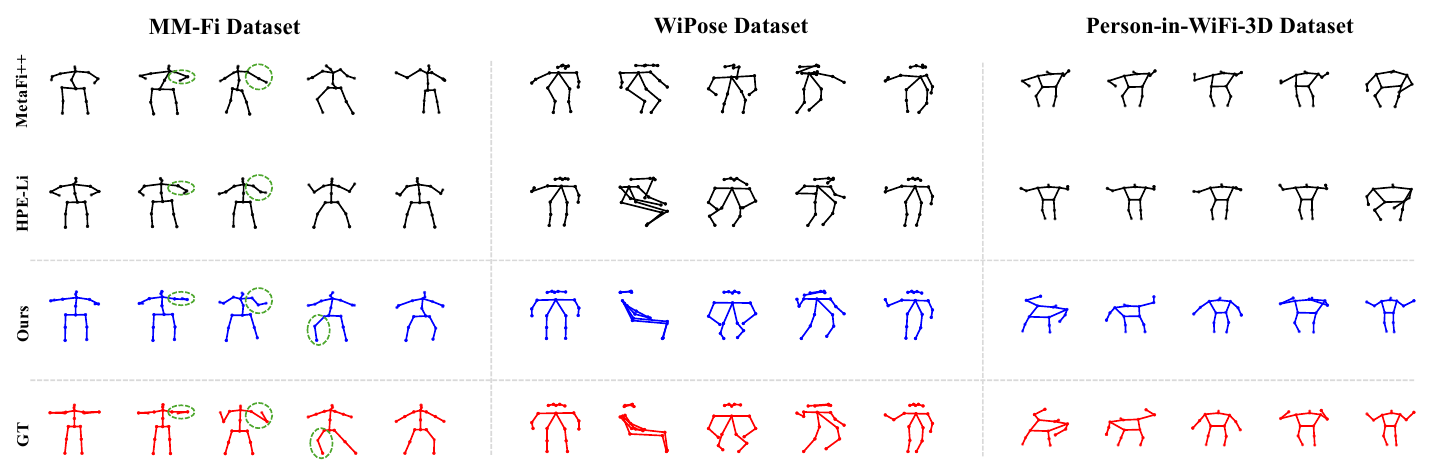} 
\end{center}
\vspace{-10pt}
\caption{Predicted poses of the MetaFi++ \cite{zhou2023metafi++}, HPE-Li \cite{d2025hpe}, and our proposed method among three different datasets.}
\label{fig:pose_comparison}
\vspace{-15pt}
\end{figure*}

\begin{table}
\caption{Performance of our method for each joint on the MM-Fi (P1-S1). L. denotes the left and R. denotes the right.}
\vspace{-8pt}
\centering
\begin{tabular}{p{1.8cm}|>{\centering\arraybackslash}p{1.7cm}>{\centering\arraybackslash}p{1.8cm}}
\hline
\rule{0pt}{10pt} Joints & MPJPE$\downarrow$ & PA-MPJPE$\downarrow$ \\
\hline
Bot Torso & 102.7 & 56.7 \\
L.Hip & 106.9 & 63.7 \\
L.Knee & 105.7 & 66.1 \\
L.Foot & 104.1 & 88.6 \\
R.Hip & 108.3 & 64.3 \\
R.Knee & 105.3 & 67.4 \\
R.Foot & 109.5 & 90.8 \\
Center Torso & 112.3 & 44.5 \\
Upper Torso & 135.8 & 54.2\\
Neck Base & 158.2 & 66.2 \\
Center Head & 160.5 & 70.9 \\
R.Shoulder & 147.7 & 73.6 \\
\rowcolor{gray!10} R.Elbow & 249.1 & 140.5\\
\rowcolor{gray!10} R.Hand & 364.5 & 284.0 \\
L.Shoulder & 141.8 & 77.2 \\
\rowcolor{gray!10} L.Elbow & 235.8 & 132.4\\
\rowcolor{gray!10} L.Hand & 362.4 & 277.0\\
\hline
Average & 165.3 & 101.0 \\
\hline
\end{tabular}
\label{tab:joints_mmfi}
\end{table}

\subsection{Qualitative Analysis}
\noindent \textbf{Masking-Reconstruction Visualization}.
In Fig. \ref{fig:ori_mask_recon}, we plot the raw, masked, and reconstructed WiFi signals, selecting four actions to highlight variations in WiFi signal patterns. Our method faithfully reconstructs the original WiFi signals, underscoring its ability to capture domain-consistent and motion-discriminative WiFi representations effectively.

\noindent \textbf{Temporal-Consistent Contrastive Learning}.
Here, we plot the t-SNE visualization of the WiFi representations, comparing models trained with and without the temporal-consistent contrastive strategy, as shown in Fig. \ref{fig:tsne-tc-cl}.  Evidently, incorporating it improves inter-sequence separability and intra-sequence compactness, thereby strengthening the temporal-consistent property within the same action sequence and prompting the motion-discriminative capacity among different action sequences.

\noindent \textbf{Dimension Collapse Phenomenon}.
In Fig. \ref{fig:dim_collpase} (a), we calculate the covariance values of each dimension of the WiFi representations. A more compact distribution is clearly visible when the uniformity term is included, implying that it enriches dimensional diversity and improves inter-dimensional dependencies. Additionally, Fig. \ref{fig:dim_collpase} (b) represents the singular values of WiFi representations. More large singular values emerge upon introducing the uniformity term, suggesting that the embedding space mitigates the dimension collapse and preserves richer information.

\noindent \textbf{Joints Analysis}.
To evaluate the joint-level accuracy of our method, we calculate the pose estimation error for each joint, as shown in Table \ref{tab:joints_mmfi}. Our method performs superior on coarse-grained body parts like the torso. However, the hands and elbows exhibit the highest errors, a finding that similarly appears in the other two datasets (Table \ref{tab:joints_wp} and Table \ref{tab:joints_piw3}) in Appendix \ref{appendix:joints}. These results stem from the limited resolution of current WiFi signals, which hinders the capture of fine-grained actions, e.g., hand movements.

\noindent \textbf{Pose Realistic.}
Fig. \ref{fig:pose_comparison} compares predicted poses across various methods and datasets. Our predictions exhibit a more consistent motion tendency in the MM-Fi dataset, highlighted by the green circles. Moreover, as the resolution of WiFi signals increases in the other two datasets, the predicted poses of our method become more coherent and precise. Notably, our predicted skeletal structures adhere closely to the human topology, with joint locations and bone lengths aligning well against the ground truth. 

\section{Conclusion}
In this paper, we revisit the task of WiFi-based human pose estimation and focus on two previously overlooked yet crucial issues: 1) cross-domain gap; and 2) structural fidelity gap. To address these challenges, we propose domain-consistent WiFi representation learning and topology-constrained pose decoding methods. In the former, we incorporate temporal-consistent contrastive learning into the self-supervised masking-reconstruction operations to learn domain-consistent and motion-discriminative WiFi representations. Additionally, uniformity regularization is employed to prevent dimension collapse. Moreover, we propose the task prompt, introduce the human topology structure into the Graph Convolution layers, and combine them with Transformer layers to explore the local-global spatial relationships among joints, constraining the decoded pose to be more realistic and fidelity. Experiments on the MM-Fi, WiPose, and Person-in-WiFi-3D datasets show the superior performance of our proposed method in 2D/3D human pose estimation tasks. 


{\small
\bibliographystyle{ieee_fullname}
\bibliography{references}
}

\clearpage
\setcounter{page}{1}
\appendix
\section{Appendix}
\subsection{Implementation Details}
In the pre-training phase, the encoder-decoder is trained for 400 epochs using the AdamW, employing a batch size of 256, a learning rate of 1.5e-4 with cosine annealing schedule, a warm-up epoch of 40, and a weight decay of 0.05. The mask ratio is set as 80\%. For the MM-Fi dataset, we train the pose decoder for 50 epochs using the SGD optimizer with a weight decay of 0.01. For the WiPose dataset, we train the pose decoder with the AdamW optimizer for 50 epochs. For the Person-in-WiFi-3D dataset, we train the pose decoder for 200 epochs using the AdamW optimizer. All the learning rates are 1e-3, and the batch size is 32. All the experiments are finished using the PyTorch platform on a GeForce RTX 4090 GPU. 

\subsection{Evaluation Metric}
Three evaluation metrics are adopted following mainstream methods \cite{yang2024mm,d2025hpe,zhou2023metafi++}:

\noindent \textbf{Mean Per Joint Position Error (MPJPE (mm))}:
Measure the average Euclidean distance between ground truth and predictions, which is widely used to evaluate absolute positional accuracy.

\noindent \textbf{Procrustes Analysis MPJPE (PA-MPJPE (mm))}:
Measure the MPJPE after aligning the predictions and ground truth using rigid transformations (translation, rotation, and scaling) by Procrustes analysis. Typically, it can be used to reflect the similarity in human shape and structure.

\noindent \textbf{Percentage of Correct Keypoints (PCK@$\alpha$ (\%))}:;
Measure the percentage of predictions that fall within a certain threshold distance from the ground truth. The threshold is set as a fraction $\alpha$ of the torso length following the previous works \cite{d2025hpe,zhou2023metafi++}. It is widely used to evaluate the local accuracy.

\subsection{WiFi Representations Comparison}
\begin{figure*}
\begin{center}
\includegraphics[width=1.0\linewidth]{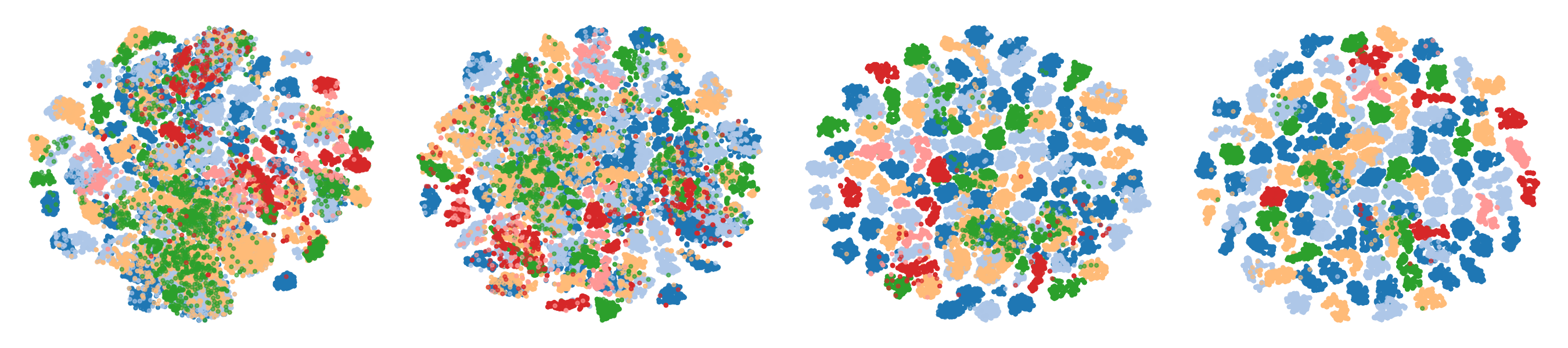}\\
\includegraphics[width=1.0\linewidth]{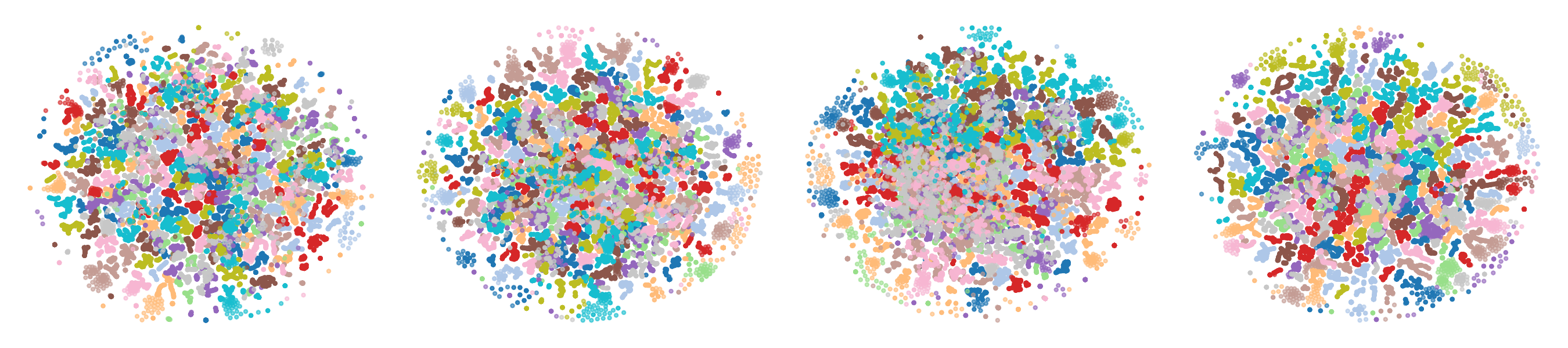}\\ \vspace{-2pt}
\begin{tabular}{@{\hskip 20pt}c@{\hskip 65pt}c@{\hskip 65pt}c@{\hskip 65pt}c@{\hskip 30pt}}
    \small{(a) Original WiFi} &
    \small{(b) HPE-Li \cite{d2025hpe}} &
    \small{(c) MetaFi++ \cite{zhou2023metafi++}} &
    \small{(d) Ours} \\
\end{tabular}
\end{center}
\vspace{-15pt}
\caption{t-SNE visualization of WiFi representations. The first row denotes the WiFi representations extracted on the MM-Fi (Protocol 1 - Setting 1) testing set, and the second row represents the representations obtained on the WiPose testing set. Each color corresponds to a distinct action category.}
\label{fig:tsne-comparison}
\vspace{-5pt}
\end{figure*}

Fig. \ref{fig:tsne-comparison} provides the t-SNE visualization of WiFi representations across multiple models and datasets. In contrast to both the raw WiFi signals and other existing methods, the learned WiFi representations of our proposed method exhibit excellent inter-sequence separability and intra-sequence compactness. Consequently, this motion-discriminative representation space benefits subsequent pose estimation tasks, whereas prior methods tend to learn spurious or motion-irrelevant features that can undermine estimation accuracy.

\subsection{Joints Analysis}
\label{appendix:joints}
As shown in Table \ref{tab:joints_wp} and Table \ref{tab:joints_piw3}, the error associated with estimating hand and elbow joints decreases when an increased number of antennas and receivers is employed for WiFi signal capture. Consequently, if the goal is to detect more fine-grained actions and reliably represent them in WiFi signals, it becomes imperative to either deploy a greater number of devices or acquire signals at higher resolutions.

\begin{table}
\caption{Performance of our method for each joint on the WiPose dataset. L. denotes the left and R. denotes the right.}
\centering
\begin{tabular}{p{1.8cm}|>{\centering\arraybackslash}p{1.7cm}>{\centering\arraybackslash}p{1.8cm}}
\hline
\rule{0pt}{10pt} Joints & MPJPE$\downarrow$ & PA-MPJPE$\downarrow$ \\
\hline
Nose & 30.9 & 14.9 \\
Neck & 27.3 & 11.7 \\
R.Shoulder & 28.7 & 13.4\\
\rowcolor{gray!10} R.Elbow & 38.7 & 25.6 \\
\rowcolor{gray!10} R.Wrist & 48.2 & 35.8\\
L.Shoulder& 29.8 & 16.6\\
\rowcolor{gray!10} L.Elbow & 37.2 & 24.9\\
\rowcolor{gray!10} L.Wrist & 43.3 & 30.6\\
R.Hip & 24.6 & 17.2\\
R.Knee & 21.0 & 19.3\\
R.Ankle & 22.6 & 21.7\\
L.Hip & 25.6 & 17.9\\
L.Knee & 22.4 & 19.2\\
L.Ankle & 26.0 & 22.2\\
R.Eye & 31.6 & 15.5\\
L.Eye & 32.4 & 16.3\\
R.Ear & 30.8 & 14.9\\
\rowcolor{gray!10} L.Ear & 96.8 & 77.7\\
\hline
Average & 34.3 & 23.1\\
\hline
\end{tabular}
\label{tab:joints_wp}
\end{table}

\begin{table}
\caption{Performance of our method for each joint on the Person-in-WiFi-3D (One-Person) dataset. L. denotes the left and R. denotes the right.}
\centering
\begin{tabular}{p{1.8cm}|>{\centering\arraybackslash}p{1.7cm}>{\centering\arraybackslash}p{1.8cm}}
\hline
\rule{0pt}{10pt} Joints & MPJPE$\downarrow$ & PA-MPJPE$\downarrow$ \\
\hline
Neck & 71.6 & 36.2 \\
Head & 77.6 & 43.1 \\
L.Shoulder & 80.5 & 37.8  \\
R.Shoulder & 81.1 & 37.8 \\
\rowcolor{gray!10} L.Elbow & 107.4 & 54.1 \\
L.Hip & 57.7 & 41.5 \\
\rowcolor{gray!10} R.Elbow & 114.2 & 55.1 \\
R.Hip & 58.6 & 42.0 \\
\rowcolor{gray!10} L.Hand & 164.8 & 117.9 \\
L.Knee & 65.6 & 52.2 \\
\rowcolor{gray!10} R.Hand & 179.2 & 122.4 \\
R.Knee & 64.3 & 52.8 \\
L.Ankle & 69.8 & 65.8 \\
R.Ankle & 67.4 & 62.5 \\
\hline
Average & 90.0 & 58.7 \\
\hline
\end{tabular}
\label{tab:joints_piw3}
\end{table}

\begin{table*}
\caption{State-of-the-art comparisons on MM-Fi dataset regarding 2D pose estimation under Protocol 3 with Setting 1 split. The best and the second-best results are marked in \textcolor{red}{\textbf{Red}} and \textcolor{blue}{\textbf{Blue}}, respectively. }
\label{table:2d_mmfi}
\centering
\begin{tabular}{l|cccccc}
\hline
\rule{0pt}{10pt} Method & PCK@20$\uparrow$ & PCK@30$\uparrow$ & PCK@40$\uparrow$ & PCK@50$\uparrow$ & MPJPE$\downarrow$ & PA-MPJPE$\downarrow$ \\
\hline
Wi-Pose \cite{jiang2020towards} & 48.6 & 65.1 & 75.6 & 82.4 & 158.2 & 97.7 \\
Wi-Mose \cite{wang2021point} & 48.7 & 66.6 & 77.3 & 83.9 & 155.8 & 95.4 \\
WiLDAR \cite{deng2023wildar} & 44.1 & 62.6 & 72.6 & 79.3 & 170.3 & 115.6 \\
WiSPPN \cite{wang2019can} & 45.4 & 63.2 & 74.1 & 81.0 & 166.5 & 110.0 \\
PerUnet \cite{zhou2022perunet} & 50.1 & 67.3 & 77.6 & 83.6 & 154.6 & 98.6 \\
MetaFi++ \cite{zhou2023metafi++} & 45.5 & 64.4 & 75.1 & 81.8 & 164.4 & 106.3 \\
HPE-Li \cite{d2025hpe} & \textcolor{blue}{\textbf{52.1}} & \textcolor{blue}{\textbf{68.2}} & \textcolor{blue}{\textbf{78.2}} & \textcolor{blue}{\textbf{85.1}} & \textcolor{blue}{\textbf{149.4}} & \textcolor{blue}{\textbf{92.5}} \\
\rowcolor{yellow!10} \textbf{DT-Pose (Ours)} & \textcolor{red}{\textbf{65.8}} & \textcolor{red}{\textbf{77.9}} & \textcolor{red}{\textbf{85.1}} & \textcolor{red}{\textbf{89.8}} & \textcolor{red}{\textbf{137.0}}  & \textcolor{red}{\textbf{92.3}} \\
\hline
\end{tabular}
\end{table*}

\end{document}